%% file: main.tex
\documentclass[final]{siamltex}
\usepackage{algorithm,algpseudocode}
\usepackage{latexsym, graphicx, epsfig, amsmath, amsfonts,amssymb,bm}
\usepackage{caption, subcaption}
\usepackage{epstopdf}
\usepackage{booktabs}
\usepackage{array}
\usepackage{color}
\usepackage{lineno}
\usepackage{url}
\usepackage{graphicx}
\usepackage{grffile}
\usepackage{subcaption}

\def\be{\begin{equation}}
\def\ee{\end{equation}}

\def\x{\mathbf{x}}
\def\y{\mathbf{y}}

\def\f{\mathbf{f}}

\def\I{\mathbf{I}}

\def\xt{\widetilde{\x}}

\def\Rs{\mathbb{R}}

\def\A{\mathbf{A}}

\def\F{\mathbf{F}}
\def\N{\mathbf{N}}
\def\Nt{\widetilde{N}}
\def\Pi{\mathbf{\Phi}}

\def\nm{n_M}

\def\F{\mathbf{F}}
\def\G{\mathbf{G}}
\def\Gt{\widetilde{\mathbf{G}}}

\title{Flow Map Learning for Unknown Dynamical Systems: Overview, Implementation, and Benchmarks}

\author{Victor Churchill\thanks{Department of Mathematics,
		Trinity College, Hartford, CT 06106, USA.
		{\tt Email: victor.churchill@trincoll.edu}.} \and Dongbin
       Xiu\thanks{Department of Mathematics,
		The Ohio State University, Columbus, OH 43210, USA.
		{\tt Email: xiu.16@osu.edu.}
		Funding: This work was partially supported by AFOSR FA9550-22-1-0011.}
}

\begin{document}
\maketitle
\begin{abstract}
Flow map learning (FML), in conjunction with deep neural networks (DNNs), has shown promises for data driven
modeling of unknown dynamical systems. A remarkable feature of FML is that it is capable of producing accurate
predictive models for partially observed systems, even when their exact mathematical models do not exist. In this paper, we present
an overview of the FML framework, along with the important computational details for its successful 
implementation. 
%These include how training data sets are prepared, as well as the detail on the DNN training. 
We also present a set of well defined benchmark
problems for learning unknown dynamical systems. All the numerical details of these problems are presented, along with their FML results, to 
ensure that the problems are accessible for cross-examination and the results are reproducible.
\end{abstract}
%\begin{keywords}
%Deep neural network
%\PACS
%\end{keywords}

% main text
\input Introduction

\input Setup
\input Method

\input Algorithms

\input Examples
%\input Conclusion
\input Appendix

\bibliographystyle{siamplain}
\bibliography{LearningEqs,neural}
%NNmemory}

%\input Appendix

\end{document}

%% file: Introduction.tex
\section{Introduction} \label{sec:intro}

Data driven modeling has received a growing amount of attention in the scientific computing community. It is concerned with the problem of learning the behavior of a system whose governing laws/equations
are unknown. The goal is to analyze and predict the system using observation data of the system.
Broadly speaking, there exist two types of approaches. The first type of approach seeks to discover, or recover, the underlying governing equations by using the data. Once the governing equations
are recovered, they can be solved numerically to analyze the system behavior. While earlier work utilized symbolic regression \cite{bongard2007automated,schmidt2009distilling}, more recent works employ
sparsity promoting numerical methods \cite{brunton2016discovering, schaeffer2017extracting, rudy2017data}. Since exact recovery of the governing equations is difficult, if not impossible, for many
practical systems, methods seeking to approximate the governing equations have also been developed. These include the use of polynomials \cite{WuXiu_JCPEQ18}, Gaussian processes \cite{raissi2017machine},
and with more popularity, deep neural networks (DNNs) \cite{raissi2018multistep,qin2018data,rudy2018deep}. The second type of approach does not seek an explicit form, exact or approximate, for the governing equations. Instead, the methods aim at constructing a numerical form to model the unknown system. Earlier works include the equation-free method \cite{kevrekidis2003equation}, and heterogeneous multiscale method (HMM) \cite{EE_HMM03}. More recent work includes dynamical model decomposition \cite{Kutz_DMDbook16}, various DNN inspired methods \cite{long2017pde,raissi2017physics1,raissi2017physics2,raissi2018deep,long2018pde,sun2019neupde}, and flow map learning method \cite{qin2018data} which is the topic of this paper.

In the study of dynamical systems, the flow map defines the mapping between the solutions at two different time instances. In flow map learning (FML), we seek a numerical approximation $\F$ to the true flow map
between two consecutive time instances by using the observation data. An accurate approximation of the flow map thus captures the dynamics of the unknown system. Subsequently, we obtain an iterative FML predictive model in the form of
$$
\x_{n+1} = \F(\x_n, \cdots).
$$
Just like a numerical time stepper, the FML model is able to conduct long-term system predictions when proper initial conditions are given. The FML approach was first proposed in \cite{qin2018data} for
fully observed autonomous systems, i.e., the observation data contain all the state variables. It was later extended to non-autonomous systems (\cite{QinCJD_Nonauto21}), where arbitrary control/excitation signals can be incorporated, and parametric systems (\cite{QinCJX_IJUQ20}) for uncertainty quantification. A major advancement was made in \cite{fu2020} for partially observed systems, where only a (small) subset of
the state variables are observed in the data. Motivated by the Mori-Zwanzig formulation  (\cite{mori1965, zwanzig1973}), the work of \cite{fu2020} extended the FML formulation by incorporating past memory terms. The resulting FML model is highly effective to produce long-term system predictions for the observed variables. Note that in this case, since the observed variables are only a subset of the entire set of the state variables (whose governing equations are unknown in the first place), exact mathematical models for the observed variables do not exist, as they depend on the other variables that are not observed. FML is particularly useful in practice, as it allows one to construct effective models for the observed variables only. 

In this paper, we provide an overview to the FML approach, for both fully observed and partially observed systems. More importantly, we provide a detailed discussion of its computational aspects, including the choices of the key parameters, stability enhancement, etc. The FML methods are particularly powerful when DNNs are used as the building blocks. We discuss the network structures, along with the technical DNN training details that are often skipped in papers.
Finally, we provide a set of well defined benchmark problems that can be replicated by users. All the computational details of the benchmark problems are provided, as well as the data used in this paper to produce the figures. This purpose is to enable users to replicate the results in the paper and provide a means of critically examining the methods.

%This paper is organized as follows. After the problem setup in
%Section \ref{sec:setup}, we present the main method in Section
%\ref{sec:method}.
%Numerical examples are then presented in 
%Section \ref{sec:examples} to
%demonstrate the properties of the proposed approach. 

%% file: Setup.tex
\section{Problem Setup} \label{sec:setup}

Let 
us consider a dynamical system, whose state variables $\y\in\Rs^n$,
$n\geq 1$, evolve according to an unknown system of governing
equations.
For example, $\y$ may follow a system of autonomous ordinary
differential equations (ODEs), 
\be \label{y}
\frac{d\y}{dt}=\f(\y),
\ee
where the right-hand-side $\f:\Rs^n\to\Rs^n$ is unknown. As a result,
the dynamics of the system can not be studied, as the governing
equations are not available for numerical simulation.

We assume that observation data for a set of {\em observables},
$\x\in\Rs^d \subseteq \y$,  $d\geq 1$, are available. Our goal is to create an accurate
predictive
model for $\x$ such that their long-term dynamics can be studied.
Two distinctly different 
cases thus exist: 
\begin{itemize}
\item Fully observed case: $\x=\y$ with $d=n$;
\item Partially observed case: $\x\subset \y$ with $d<n$.
  \end{itemize}

  Note that our discussion in this paper is not restricted to systems
  of ODEs like \eqref{y}. The unknown governing
equations may take rather complicated forms and include various
(algebraic) constraints.

\subsection{Observation Data}

Let us assume the observation data are time sequences of the
observables $\x$, in the following form,
%
%For notational convenience, let us assume the observations are made
%over time instances of a constant time step 
%
%\be \label{X}
%\X^{(i)} = \left\{ \x\left(t^{(i)}_k\right)\right\}, \qquad k=1,\dots, K^{(i)},
% \quad i=1,\dots, N_T,
% \ee
 \be \label{Xt}
 \begin{split}
   &%\left\{
     \x\left(t^{(1)}_0\right), \cdots,
     \x\left(t^{(1)}_{L_1}\right);
     % \right\},
     \\
     & %\left\{
       \x\left(t^{(2)}_0\right), \cdots,
       \x\left(t^{(2)}_{L_2}\right);
     %\right\},
     \\
  & \cdots \cdots\\
  & %\left\{
  \x\left(t^{(\Nt)}_0\right), \cdots,
  \x\left(t^{(\Nt)}_{L_{\Nt}}\right).
%\right\}.
 \end{split}
 \ee
 Each sequence represents an evolution {\em trajectory} data of $\x$, and
 $\Nt\geq 1$ is the total number of such trajectories.
 For simplicity, we assume the data are over evenly
distributed time
instances with a uniform time step $\Delta$, i.e.,
\be \label{Delta}
\Delta \equiv t_{k}^{(i)} - t_{k-1}^{(i)}, \qquad \forall k, i.
\ee
Note that each of the $i$-th trajectory, $i=1,\dots,\Nt$, has its own
``initial condition'' $\x(t^{(i)}_0)$ that may occur at
different time $t^{(i)}_0$.

\subsection{Modeling objective}

Our goal is to create a numerical time marching model
\be\label{G}
\x_{n+1} = \mathbf{G}(\x_n, \dots),
\ee
such that for any given proper initial conditions on $\x $, the model
prediction is an accurate approximation to the true solution, i.e.,
\be
\x_n \approx \x(t_n), \qquad n=0,\dots,
\ee
where we require $n\gg 1$, as
long-term prediction is our primary focus.

For the fully observed case when $\x=\y $, the model \eqref{G} can be
considered
a numerical approximation for the unknown governing equation in
\eqref{y}.
For the partially observed case when $\x\subset \y$, an exact mathematical
model for $\x $ often does not exist. This challenging case is more
practical, as in practice it is often difficult, if not impossible, to
acquire observation data on all the components of the system variables
$\y\in\Rs^n $. Often, one only has data on a small number of variables. This
implies $\x\in\Rs^d\subset \y$, with $d\ll n$.

%% file: Method.tex
\section{Flow Map Learning Methods} \label{sec:method}

In this section, we overview the key ingredients of the FML
method. We first discuss the need to avoid using the time variable in the
process, and then discuss the FML modeling approach and its
corresponding numerical components.

\subsection{Key principle: time variable removal} \label{sec:principle}

An important design principle of FML is that the time variable shall not
explicitly present itself in the process. This implies that in FML, we
shall not seek to construct a numerical approximation to $\x(t)$ as a
function of time. To understand this principle, we consider the
contrary. Suppose our modeling method requires an approximation
$\xt (t)\approx \x(t)$. Then, such an approximation needs to be constructed using the
data set \eqref{Xt}. Let $[T_L, T_U]$ be the smallest time domain
containing all the time instances in the data set
\eqref{Xt}, 
$$
0\leq T_L \leq \min_{k,i}  t_{k}^{(i)} < \max_{k,i}  t_{k}^{(i)} \leq
T_U < \infty.
$$
Naturally, an accurate approximation $\xt(t)$ can
only stay accurate within this training time domain, i.e., $\xt(t) \approx \x(t)$,
$t\in [T_L, T_U]$. However, for long-term prediction of the dynamics
of $\x(t)$, we need $t\gg T_U$. Consequently, $\xt(t)$ becomes an
extrapolating approximation of $\x(t)$ (far) beyond the training time
domain. It is well known in approximation theory that extrapolation is
a highly unstable procedure. No matter what approximation method is
used inside the interval $[T_L, T_U]$, numerical accuracy can not be
ensured outside the interval
$[T_L, T_U]$, especially far beyond it.
Therefore, if a temporal
approximation $\xt(t)$ is sought after in a learning method, the
method will not be able to provide 
accurate predictions beyond the training time domain $[T_L,
T_U]$. This does not serve our purpose.

By following this design principle, we first eliminate the time
variables in the trajectory data \eqref{Xt} and rewrite into the following:
 \be \label{X}
 \begin{split}
   & %\left\{
   \x^{(1)}_0, \cdots,
   \x^{(1)}_{L_1}
 %\right\}
 ;\\
 & %\left\{
   \x^{(2)}_0, \cdots,
   \x^{(2)}_{L_2}
   % \right\}
   ;\\
  & \cdots \cdots\\
  & %\left\{
  \x^{(\Nt)}_0, \cdots,
  \x^{(\Nt)}_{L_{\Nt}}
%\right\}
.
 \end{split}
 \ee
The difference between \eqref{X} and \eqref{Xt}, albeit subtle, is
significant. In the data set \eqref{X}, the absolute time $t$ is not
present, and only the relative time shifts
among different data entries matter. Hereafter, we shall refer to
\eqref{X} as the {\em raw data set}. The use of \eqref{X} also eases
the practical burden of data acquisition, as one no longer needs to
record the time variable.

\subsection{FML model}

The general form of the FML model, developed in \cite{qin2018data,
  fu2020}, takes the following form:
 \be \label{FML}
 \x_{n+1} = \mathbf{G}(\x_n, \dots, \x_{n-\nm}), \qquad n\geq \nm\geq 0,
 \ee
 with initial conditions supplied for
 $
\x_0, \dots, \x_{\nm}.
$
Here, $\nm\geq 0$ is an integer called the {\em memory step}. For partially observed systems
with $\x\subset \y$, accurate modeling of the observable $\x$ requires
a certain length of memory $T_M>0$. (See \cite{fu2020} for the
derivation.) The memory step $\nm$ is the number of time steps
within the memory $T_M$, i.e., $\nm \Delta =
T_M$. Hence, $\nm >0$ for partially observed systems. The precise
value of $\nm$, or equivalently $T_M$,  is
problem dependent. See the discussion in Section \ref{sec:MK}
below. For fully observed system $\x=\y$, $\nm =0$.

It is often preferred, from a computational point of view, to rewrite
the FML in the following equivalent form,
 \be \label{FML1}
 \x_{n+1} = \x_n + \Gt(\x_n, \dots, \x_{n-\nm}), \qquad n\geq \nm\geq 0,
 \ee
 where $\Gt = \G - \I d$ with $\I d$ being the identity operator. In
 this form, the FML model resembles a multi-step time integrator.

\subsubsection{FML for fully observed system}

For fully observed system $\x=\y$, there is no need for memory
terms. Subsequently, $\nm=0$.
The FML model takes the following reduced form
\be \label{Euler}
\x_{n+1} = \mathbf{G}(\x_n), \qquad n\geq 0.
\ee
with the initial condition specified for $\x_0$. This can also
be written in the following equivalent form
 \be \label{Euler1}
 \x_{n+1} = \x_n + \Gt(\x_n), \qquad \Gt = \G - \I d.
 \ee
 In this form, it resembles the Euler forward time integrator. As pointed out
 in \cite{qin2018data}, however, this is not the Euler forward integrator.

 The goal is then to construct $\G$ in \eqref{Euler}, or,
 equivalently, $\Gt$ in \eqref{Euler1}, by using the data set
 \eqref{X}.
 Upon realizing that $\G$ in \eqref{Euler} serves as a mapping from
 $\x_n$ to $\x_{n+1}$, we take any two consecutive data entries from
 the raw data
 set \eqref{X} to enforce this mapping. By doing so, we construct the
 following
 training data set consisting of data pairs
\be \label{full_train}
%\X^{(i)} =
\left\{ \x^{(i)}_0,
   \x^{(i)}_1\right\}, \qquad i=1,\dots, N,
 \ee
 where $\x^{(i)}_0$ is any entry in the raw data set \eqref{X}, 
 $\x^{(i)}_1$ is the next entry immediately following $\x^{(i)}_0$, and $N\geq
 1$ is the total number of such pairs chosen from \eqref{X}. We use
 the subscript 0 and 1 to emphasize that each of these data pairs can
 be considered a trajectory of $\x$ starting with an ``initial
 condition'' $\x^{(i)}_0$ and marched forward by a single time step to
 reach $\x^{(i)}_1$. Note
 that it is possible to choose multiple such data pairs from any
 single trajectory in the raw data set \eqref{X}.

 The problem of finding $\G$ can then be formulated as follows. Given
 the training data set \eqref{full_train}, 
 find $\G:\Rs^d\to\Rs^d$ such that the mean-squared loss (MSE)
 \be \label{loss0}
 \frac{1}{N}\sum_{i=1}^N \left\|\x_1^{(i)} -
   \G\left(\x_0^{(i)}\right)\right\|^2
 \ee
 is minimized. In practice, $\G$ is written in a parameterized form
 $\G = \G(\cdot; \Theta)$, where $\Theta\in\Rs^{n_\Theta}$,
 $n_\Theta\geq 1$, are the
 hyper parameters. The optimization problem becomes
  \be \label{min0}
 \min_\Theta \frac{1}{N}\sum_{i=1}^N \left\|\x_1^{(i)} -
   \G\left(\x_0^{(i)}; \Theta\right)\right\|^2.
 \ee
 The parameterization of $\G$ can be accomplished by any proper form,
 e.g., orthogonal polynomials, Gaussian process (GP), DNNs, etc.
 
\subsubsection{FML for partially observed system}

For partially observed systems with $\x\subset \y$, there usually does
not exist a mathematical model for the evolution of $\x$. Motivated by
Mori-Zwanzig formulation, the work of \cite{fu2020} derived that the
evolution of the dynamics of $\x\in\Rs^d$ at any time $t_n$ follows
\be
\left.\frac{d\x}{dt}\right|_{t_n} = \F(\x_n, \dots, \x_{n-\nm}),
\ee
where $\nm\Delta = T_M>0$ is the memory length for the system, and
$\F$ is an unknown operator. The FML model thus follows
 \be 
 \x_{n+1} = \mathbf{G}(\x_n, \dots, \x_{n-\nm}), \qquad n\geq \nm> 0,
 \ee
 where $\G:\Rs^{d\times (\nm+1)}\to\Rs^d$ is unknown. In order to
 learn $\G$, we choose trajectories of length $(\nm+2)$ from the raw data
 set \eqref{X}.
 Our training data set thus consists of such ``trajectories'' of length $(\nm+2)$,
 \be \label{partial_train}
 %\X^{(i)} =
 \left\{ \x^{(i)}_0,\dots,
   \x^{(i)}_{\nm+1}\right\}, \qquad i=1,\dots, N,
 \ee
 where $\x^{(i)}_0$ is any data entry in the raw data set \eqref{X}, and $\x^{(i)}_1$
 to $\x^{(i)}_{\nm+1}$ are the $(\nm+1)$ entries immediately
 following. This requires that the length of each trajectory in the raw data set
 \eqref{X} to be sufficiently long, $L_i^{(i)}\geq \nm+1$,  $\forall i$. If
 any 
 trajectories in \eqref{X} are shorter than this, they shall be
 discarded. If a trajectory in \eqref{X} has a longer length, $L_i^{(i)}> \nm+1$,
 it is then possible to choose multiple $(\nm+2)$-length trajectories
 for the training data set \eqref{partial_train}.

The problem of constructing the
FML model \eqref{FML} becomes, given the training data set \eqref{partial_train},
 find $\G:\Rs^{d\times (\nm+1)}\to\Rs^d$ such that the mean-squared loss (MSE)
 \be \label{loss1}
 \frac{1}{N}\sum_{i=1}^N \left\|\x_{\nm+1}^{(i)} -
   \G\left(\x_0^{(i)},\dots, \x_{\nm}^{(i)}\right)\right\|^2
 \ee
 is minimized.
 Upon expressing $\G$ in a parameterized form $\G(\cdots; \Theta)$,
 the minimization problem becomes
  \be \label{min1}
 \min_\Theta \frac{1}{N}\sum_{i=1}^N \left\|\x_{\nm+1}^{(i)} -
   \G\left(\x_0^{(i)},\dots, \x_{\nm}^{(i)}; \Theta\right)\right\|^2.
 \ee
 Note that the fully observed case \eqref{min0} is a special case of $\nm=0$.

 \subsubsection{Function Approximation and DNNs}

 It is important to recognize that although $\G$ in FML
 \eqref{FML}  is an evolution operator governing the evolution of  the state variable
 $\x(t)$, FML is not an operator learning method. Instead, in FML \eqref{loss1},
 resp. \eqref{loss0}, $\G$ is sought after as a multivariate function
 approximation problem. This is a direct consequence of the principle
 discussed in Section \ref{sec:principle}, as the time variable $t$
 does not explicitly enter the FML formulation. This is also the main
 reason FML models can conduct accurate long-term predictions of $\x$.

\begin{figure}[htbp]
	\centering
{\includegraphics[width=0.6\textwidth]{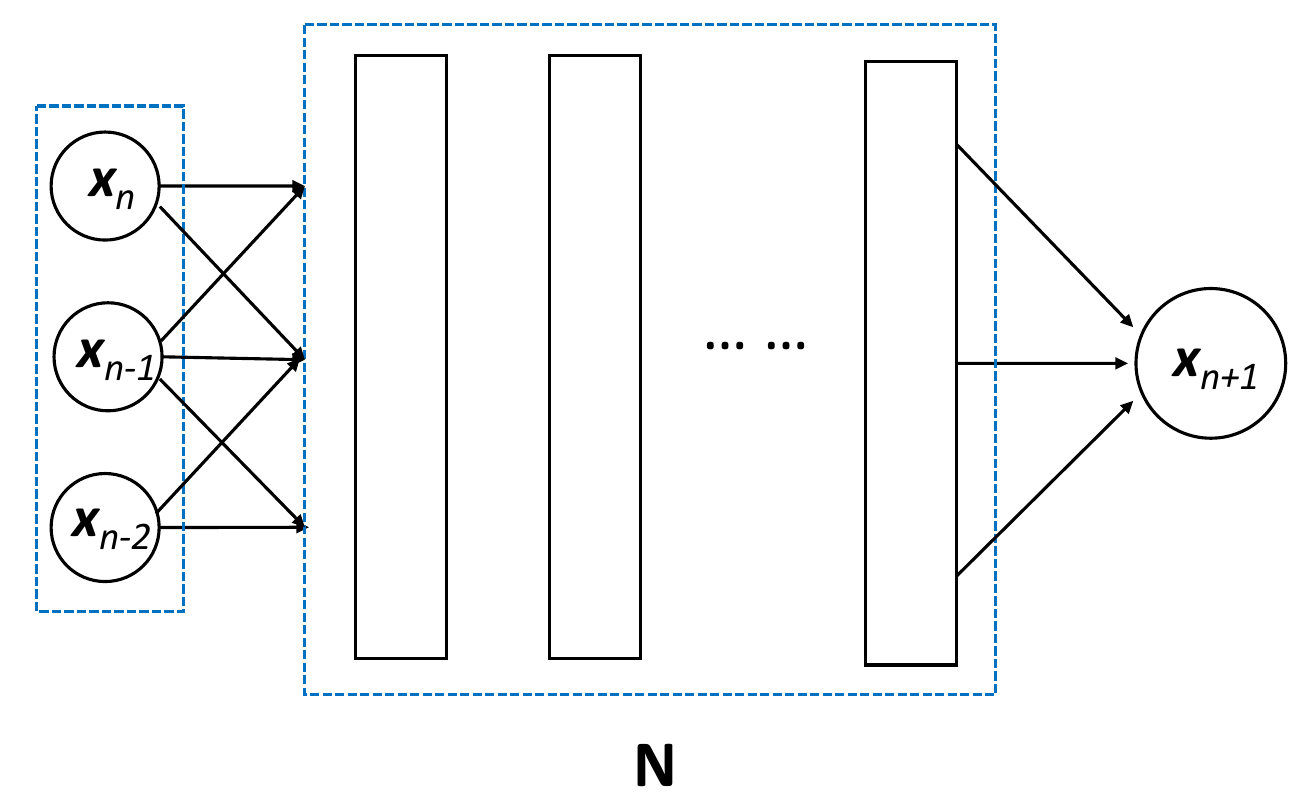}}
	\caption{Basic DNN structure for FML with $\nm=2$.}
              \label{fig:MZ}
            \end{figure}

 To numerically approximate $\G:\Rs^{d\times (\nm+1)}\to\Rs^d$, any
 multivariate function approximation methods can be adopted.
 The input dimension of $\G$ is $n_G = d\times (\nm+1)$. If
 $n_G$ is small, orthogonal polynomials are perhaps among the best
 options (\cite{WuXiu_JCPEQ18}).
 In many cases, $n_G$ is likely to be large, then DNNs become
 a more viable option. Let $\N(\cdot; \Theta)$ be the DNN mapping
 operator, where $\Theta$ stands for the hyper parameters (weights,
 biases) of the network. The DNN training is thus conducted via
 \eqref{min1} for the network parameters. The DNN structure for FML is
 illustrated in Figure \ref{fig:MZ}. Note that there is no need to
 employ any special network structures. A plain feedforward fully
 connected network is sufficient. The special case of fully observed
 system with $\nm=0$ is self explanatory. We remark that the effect of
 the memory terms can also be realized via the use of the well known LSTM
(long-short term memory) networks. The derivation for FML here (see
\cite{fu2020} for details) thus makes it clear that the memory is a
mathematical necessity and its realization can be accomplished by the
simple DNN in Figure \ref{fig:MZ}.
 %%%%%%%%%%%%%%%%%
            %%%%%%%%%%%%%%%%%%%%%%%%%

%% file: Algorithms.tex
\section{Implementation Details}

In this section, we discuss the computational details to construct
accurate FML models. Most of the details are associated with the use
of DNNs in the FML formulation.

\subsection{Multi-step loss}

The long-term stability of the FML model \eqref{FML} is not fully
understood, largely due to our lack of understanding of the
fundamental properties of DNNs. After extensive numerical
experimentations, we have come to the conclusion that the stability of
the FML models can be significantly enhanced by adopting a multi-step
loss function during the training process.

The idea of multi-step loss is to take the averaged loss of the FML
model over several time steps ahead. Let $K\geq 0$ be the number of
time steps ahead. Our training data set now consists of trajectories
$K$ steps longer than those in \eqref{partial_train}:
 \be \label{training_data}
 %\X^{(i)} =
 \left\{ \x^{(i)}_0,\dots,
   \x^{(i)}_{\nm+1}, \dots,  \x^{(i)}_{\nm+1+K}\right\}, \qquad K\geq
 0, \quad i=1,\dots, N.
 \ee
 For each of the $i$-th training trajectory, $i=1,\dots, N$, we use $\x_0^{(i)},
 \dots, \x_{\nm}^{(i)}$ as initial conditions and conduct the system
 predictions by the FML model for ($K+1$) steps:
% 
 %$\xt_0^{(i)} = \x_0^{(i)}, \dots, \xt_{\nm}^{(i)} = \x_{\nm}^{(i)}$
 \begin{equation*}
   \begin{split}
&     \xt_0^{(i)} = \x_0^{(i)}, \quad \dots, \quad \xt_{\nm}^{(i)} =
\x_{\nm}^{(i)}, \\
& \xt_{n+1}^{(i)} =  \G\left(\xt_n^{(i)},\dots, \xt_{n-\nm}^{(i)}\right), \qquad
n=\nm, \dots, \nm+K.
\end{split}
\end{equation*}
The averaged loss
 against $\x_{\nm+1}^{(i)}, \dots, \x_{\nm+1+K}^{(i)}$ are then
 computed,
 \be \label{multi_loss}
 \frac{1}{N(K+1)}\sum_{i=1}^N \sum_{k=0}^K\left\|\x_{\nm+1+k}^{(i)} -
   \xt_{\nm+1+k}^{(i)}\right\|^2.
 \ee
 The FML operator $\G$ is then computed by minimizing this multi-step loss function.
 It is obvious that the loss function \eqref{loss1} is a special case
 of $K=0$.

 \subsection{Choices of Key parameters} \label{sec:MK}

 There are three key parameters in the FML modeling process: the time step
 $\Delta$, the number of memory step $\nm$, and the number of multi-step loss $K$.

 \begin{itemize}
 \item {\em Time step} $\Delta$. In FML modeling, the choice of the
   size of the time step $\Delta$ is made primarily based on the need
   for proper temporal resolution. In most cases, if one has the
   freedom to choose the time step in the raw data set \eqref{X}, it
   is preferred to have it reasonably small to resolve the
   dynamics. The impact of $\Delta$ on the long-term stability of the
   FML prediction is not well understood. Our extensive numerical
   experimentation has indicated that FML models using DNNs are quite
   stable, when multi-step loss $K>0$ is used and the DNNs are trained
   with sufficient accuracy.

\item {\em Multi-step loss} $K$. Using multi-step loss with $K>0$ can
  significantly improve the long-term numerical stability of the
  trained FML model. 
  Based on our extensive numerical testing, we have
  found that $K=5\sim 10$ is a balanced choice.
Too small a $K=1\sim 3$ results in unreliable performance for
enhancing stability, whereas too big a $K>10$ does not offer any
noticeable gain while significantly increasing
the computational cost for DNN training.

 \item {\em Memory step} $\nm$. The choice of the memory step $\nm$ is
   perhaps the most important one. For fully observed system $\x=\y$,
   it is an easy choice to set $\nm=0$. For partially observed system
   $\x\subset \y$, the choice is less obvious, as it depends on the
   effective memory $T_M$ of the system in terms of  $T_M = \nm
   \Delta$. To determine $\nm$ or, equivalently, $T_M$, we adopt
  a ``resolution independence'' procedure. That is, we start
   with a small $\nm$,  and train several FML models using
   progressively larger $\nm$. Upon comparing the long-term
   predictions of the models, we stop the procedure when the FML
   model predictions start to converge. The corresponding
   $\nm$ and $T_M$ are then determined.
   It is important to recognize
   that although a large $T_M$ is always preferred mathematically, as
   indicated by the derivation in \cite{fu2020}, using too large a
   $T_M$ not only makes the FML modeling computational expensive, it
   also induces numerical instability for the trained FML model. This is due to
   error accumulation of the terms in the ``far
   back'' of the memory. Those terms should have (almost) zero memory effect but
   numerically they are never zero due to training errors.

 \end{itemize}
 
 \subsection{Training Data Generation}

Once the number of memory steps $\nm$ and multi-step loss $K$ are
determined, the total number of data entries required to compute
the loss function \eqref{multi_loss} is
\be \label{n_data}
n_{data} = \nm + K +2.
\ee
Thus, each trajectories in the raw data set \eqref{X} needs to be
sufficiently long such that $L_i>n_{data}$, $\forall i$. The trajectories with shorter
length are then discarded. For trajectories with length longer than
$n_{data}$, it is possible to choose multiple segments of
$n_{data}$ consecutive data entries to form the training data set:
 \be \label{final_data}
\left\{ \x^{(i)}_0,\dots,
   \x^{(i)}_{\nm}, \dots, \x^{(i)}_{\nm+1+K}\right\}, \qquad i=1,\dots, N.
 \ee

Sometimes, one has the luxury to decide how to collect the original raw data set
\eqref{X}. 
It is then highly beneficial to choose the trajectories
more strategically. Realizing that the FML model \eqref{FML} is a
function mapping $\G:\Rs^{d\times (\nm+1)}\to\Rs^d$, we can focus on
sampling the observables $\x\in \Rs^d$. To this end, the ``large number
of short bursts'' sampling strategy proposed in \cite{WuXiu_JCPEQ18} is
highly effective.

Let $\Omega_X$ be the phase
space where we can interested in the dynamics of $\x$.
%That is,
%$\x\in\Omega_X\subset\Rs^d$.
Without loss of
much generality, we assume it is a bounded hypercube. Specifically,
$x_i \in [a_i, b_i]$, where $b_i>a_i$, $i=1,\dots, d$, and $\x\in
\Omega_X= \otimes_{i=1}^d [a_i, b_i]\subset  \Rs^d$.
%$\Omega_X = \otimes_{i=1}^d [a_i, b_i]$, where $b_i>a_i$. In another
%word, each component of $\x$ is in bounded interval,
Our training data generation consists the
following two steps:
\begin{itemize}
  \item {\em Generating raw data set \eqref{X}.} We first generate
    $\Nt$ trajectory data in \eqref{X}. For notational simplicity, let
    each trajectory of the same length, i.e., $L\equiv L_i,$
    $i=1,\dots, \Nt$. We first sample $\Nt$ ``initial conditions''
    $\x_0^{(i)}$, $i=1,\dots, \Nt$, in the domain $\Omega_X$. We
    advocate random sampling with a uniform distribution in
    $\Omega_X$, unless other prior knowledge of the distribution of
    $\x$ is available. For each initial condition $\x_0^{(i)}$, we
    then collect its subsequent state to form a length $L>n_{data}$
    trajectory. In fact, we advocate the use of slightly longer
    $L$. Let $L=\gamma\cdot n_{data}$, with $\gamma=2\sim 10$. The use
    of the longer $L$ is to let the trajectory to start to converge to its true
    distribution in the phase space $\Omega_X$. 
    
  \item {\em Trajectory subsampling}. With each trajectory in
    \eqref{X} longer than the required length in the training data
    set, i.e. $L>n_{data}$, we then randomly choose, with a uniform
    distribution, $n_{burst}\geq 1$ segments of $n_{data}$ consecutive
    entries to include in the training data set
    \eqref{final_data}. The total number of training trajectories in
    \eqref{final_data} is $N=n_{burst} \cdot \Nt$. Note that when the
    raw data set \eqref{X} is given, using a larger number of
    $n_{burst}$ can boost the size of the training data set
    \eqref{final_data}.
    However, since each
    trajectory in the raw data set \eqref{X} gives $n_{burst}$ trajectories of length
    $n_{data}$, they represent a group of highly clustered entries in
    the domain $\Omega_X$. This is undesirable for accurate function
    approximation. Therefore, we strongly advocate to keep
    $n_{burst}<10$, and whenever possible, use a smaller
    value. Ideally, when the raw data set \eqref{X} is easy to
    generate, we prefer to keep
    $n_{burst}=1$ and subsequently, $N=\Nt$.

  \end{itemize}

\subsection{DNN Training}

For low-dimensional systems when $\G$ can be approximated by the
standard methods such as orthogonal polynomials, there are many fewer
uncertainties associated with the learned $\G$. However, when a DNN is
used to represent $\G$, its learning induces much larger uncertainties due to
the inherent randomness in the DNN training.

In order to obtain accurate and reproducible results, we present the
following ``guidelines'' that are based on our extensive DNN training
experience in the context of FML.
\begin{itemize}
  \item {\em Training loss.} It is essential that the training loss decays
    at least 2 orders of magnitude from its value at the beginning of
    the training. This is usually accomplished via training over a
    large number of epochs. In most of our computations, the results
    were obtained with 10,000 to 20,000 epochs. Sometimes even
    more. To examine the training loss decay, log-scale plotting is
    also required.
    \item {\em Validation loss.} It is customary to use a small
      percentage, typically 5\% to 10\%, of the training data to
      compute validation loss. However, for FML, the goal is to
      achieve highest possible accuracy in the entire phase space of
      $\x$. A small number of samples by the validation set are not
      sufficient to give a good indication of the error
      distribution. Therefore, we do not advocate the use of
      validation set in FML training. 
    \item {\em Optimization and Learning rate.} We have found the Adam
      optimizer to be sufficient in FML model training. The default
      learning rate of $10^{-3}$ in most DNN packages 
      is usually too big. We suggest the use of a smaller learning
      rate on the order of
      $10^{-4}$, or cyclic training with $10^{-3}$ as the upper
      bound. Whenever the training loss does not decay at the early
      stage of the training, it is an indication that a smaller
      learning rate is necessary.
    \item {\em Best model.} Saving the best model based on training
      loss, not validation loss, is important.
    \item {\em Ensemble averaging.} This is perhaps the most effective
      way to
      enhance reproducibility of the FML models (\cite{Churchill_Ensemble23}). With everything else
      fixed, we vary the initial random seeds of the optimizer and
      train $N_{model}>1$ DNN models independently. This in turns
      gives $N_{model}$ independent FML predictive models.
      During system
      prediction, each model starts with the same initial conditions
      and marchs forward one time step. The results are then averaged
      and served as the same new initial conditions to all the models in
      the next time step. (Note this is fundamentally different from
      averaging the long-term predictions of all models.) We have found that
      $N_{model}= 5\sim 10$ is highly effective to reduce the
      prediction randomness in practice. Details and derivation of
      ensemble averaging can be found in \cite{Churchill_Ensemble23}.
      
      \end{itemize}

%% file: Examples.tex
%\documentclass{article}
%\usepackage{graphicx}
%\begin{document}

\section{Numerical Benchmarks} \label{sec:examples}

In this section, we present a set of well defined benchmark problems
for modeling unknown dynamical systems. The problems include small
linear systems, nonlinear chaotic systems, as well as a larger linear
system. Both fully observed and partially observed cases are
considered. To summarize, the key parameters reported in each example include
\begin{itemize}
    \item $\Delta$: time step.
  \item $\Omega_X$: domain for initial condition sampling.
  \item $\Nt$: number of trajectories in the raw data set \eqref{X}.
    \item $L$: length of the trajectories in \eqref{X}.
    \item $N_{burst}$: number of training data trajectories taken from
      each raw data trajectories.
      \item $N$: total number of training trajectories in the training
        data set \eqref{final_data}:
        $N=n_{burst}\cdot \Nt$.
                \item $\nm$: number of memory steps.
          \item $K$: number of multi-step loss.
        \item DNN structure: number of hidden layers, number of nodes
          per layer.
        \item Learning rate and the number of epochs for DNN training.
          \item $n_{model}$: number of independent FML models trained
            for ensemble averaging.
      \end{itemize}

\subsection{Small decaying linear system}

The first problem is a small $2\times 2$ linear system:
\begin{equation}
\frac{d\mathbf{y}}{dt} = \A\mathbf{y},\qquad \A=\begin{bmatrix} 1 & -4 \\ 4 & -7 \end{bmatrix}.
\end{equation}
The exact solution, which decays over time, is readily available to serve as the reference
solution.

\subsubsection{Fully observed case}

%First we model the full system after observing both variables. We select $10^4$ initial conditions from $U[0,2]^2$ and solve the system numerically with $\Delta t = 10^{-2}$ for $t\in[0,2]$.
%$5$ bursts are chosen from the $10^4$ training trajectories, yielding
%$5\times10^4$ total training samples.

The key parameters are:
\begin{itemize}
    \item $\Delta=0.01$.
  \item $\Omega_X=[0,2]^2$.
  \item $\Nt=10^4$.
    \item $L=200$.
    \item $N_{burst}=5$
      \item 
        $N=n_{burst}\cdot \Nt = 5\times 10^4$.
                \item $\nm=0$
          \item $K=10$.
        \item DNN structure: 3 hidden layers with 10 nodes
          per layer.
        \item Learning rate is $10^{-4}$; number of epochs is $10^4$.
          \item $n_{model} = 10$.
      \end{itemize}

\begin{figure}[htbp]
	\begin{center}
		\includegraphics[width=\textwidth]{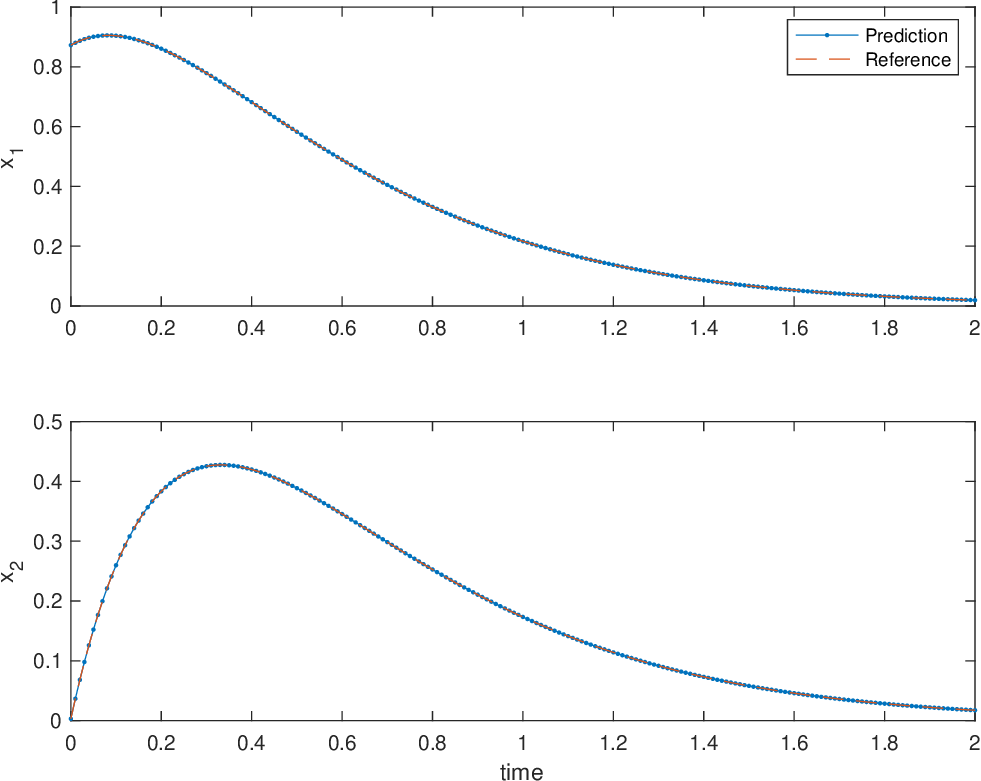}
		\caption{Benchmark 5.1.1: Example trajectory prediction.}
		\label{fig:smalllinear_full_example}
	\end{center}
\end{figure}

\begin{figure}[htbp]
	\begin{center}
		\includegraphics[width=0.8\textwidth]{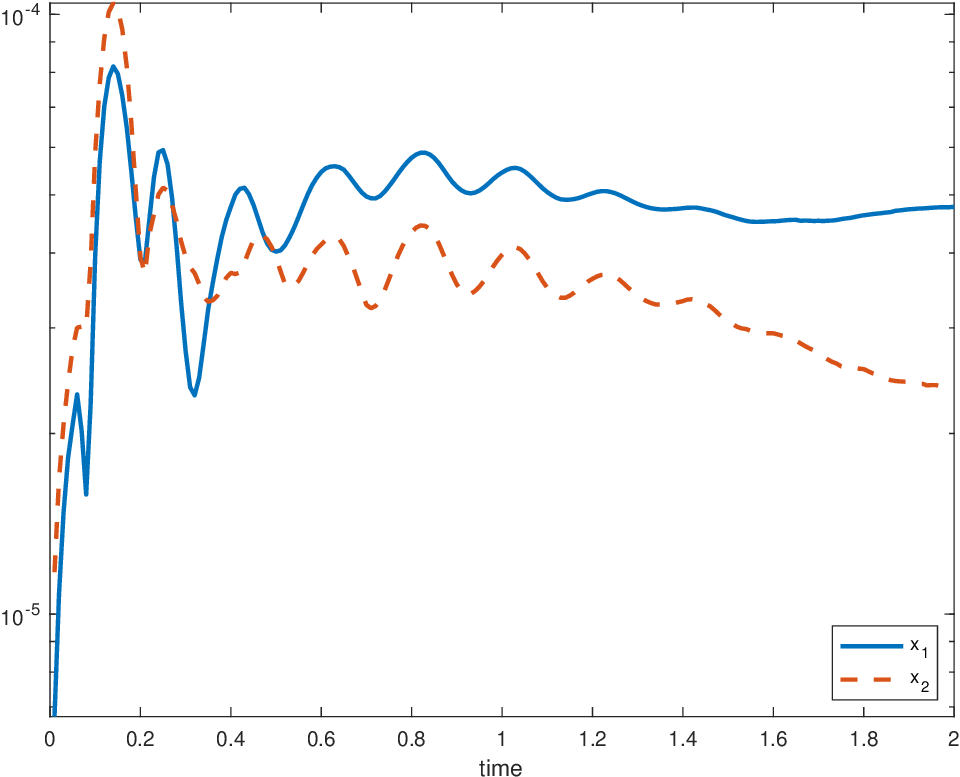}
		\caption{Benchmark 5.1.1: Average trajectory prediction.}
		\label{fig:smalllinear_full_average}
	\end{center}
\end{figure}

      For validation testing, we use $10^2$ new initial conditions in
      $\Omega_X$.
%A $10$-model ensemble is used, and each network inputs a $1$-step
%memory period and has $3$ hidden layers each with $10$ hidden
%nodes. The $10$-step recurrent loss is trained for $10^4$ epochs with
%a learning rate of $10^{-4}$.
Figure \ref{fig:smalllinear_full_example} shows an example trajectory,
and Figure \ref{fig:smalllinear_full_average} shows the average
$\ell_2$ error over the 100 test trajectories.

\subsubsection{Partially observed case}

We set the observable to be $x = y_1$. Obviously, the
training data set \eqref{final_data} contains only the trajectory data
of $y_1$.
The memory steps, after
testing, is chosen to be $\nm=10$. The rest of the key parameters are
the same as in the previous example.
The key parameters are:
\begin{itemize}
    \item $\Delta=0.01$.
  \item $\Omega_X=[0,2]^2$.
  \item $\Nt=10^4$.
    \item $L=200$.
    \item $N_{burst}=5$
      \item 
        $N=n_{burst}\cdot \Nt = 5\times 10^4$.
                \item $\nm=10$
          \item $K=10$.
        \item DNN structure: 3 hidden layers with 10 nodes
          per layer.
        \item Learning rate is $10^{-4}$; number of epochs is $10^4$.
          \item $n_{model} = 10$.
      \end{itemize}

%All the parameters are the same
%as in the previous example, except that $\nm=10$. 
%We also can model the reduced system after observing only the first variable. We select $10^4$ initial conditions from $U[0,2]^2$ and solve the system numerically with $\Delta t = 10^{-2}$ for $t\in[0,2]$.
%$5$ bursts are chosen from the $10^4$ training trajectories for only the first variable, yielding $5\times10^4$ total training samples. $10^2$ new test initial conditions for the first variable are also collected from $U[0,2]$.

\begin{figure}[htbp]
	\begin{center}
		\includegraphics[width=0.8\textwidth]{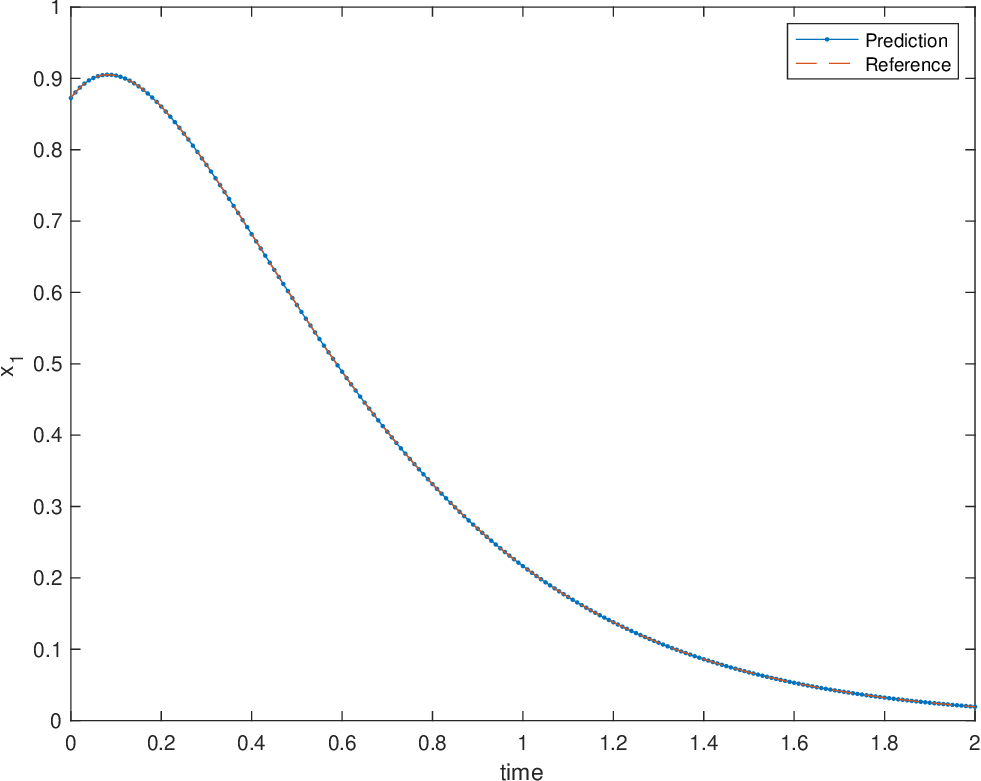}
		\caption{Benchmark 5.1.2: Example trajectory prediction.}
		\label{fig:smalllinear_reduced_example}
	\end{center}
\end{figure}

\begin{figure}[htbp]
	\begin{center}
		\includegraphics[width=0.8\textwidth]{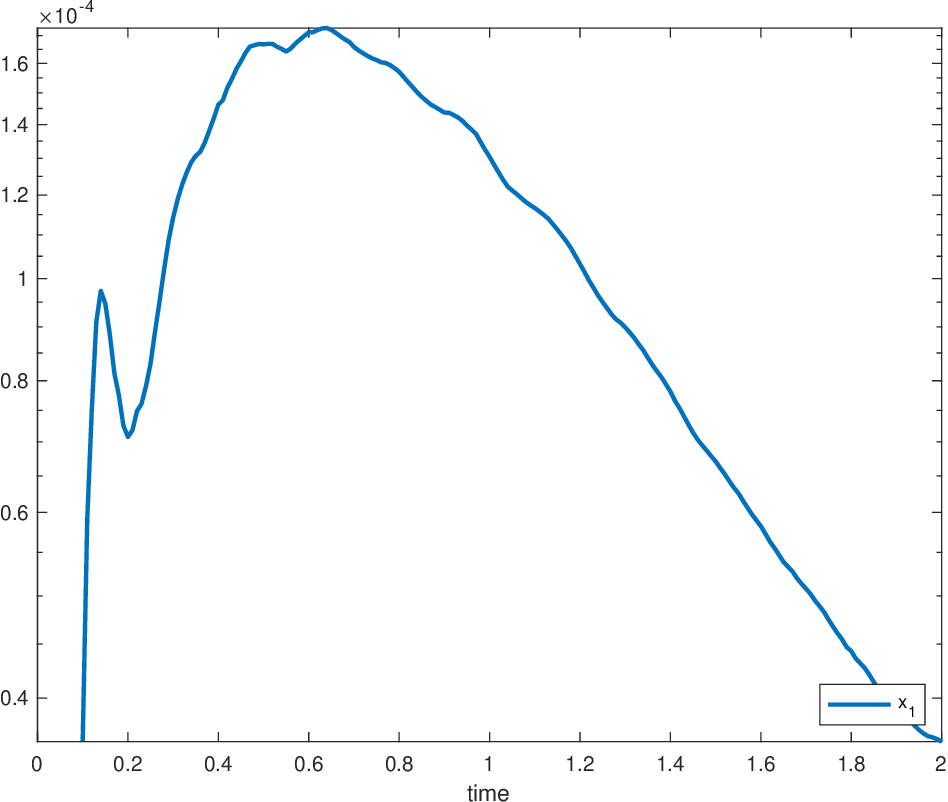}
		\caption{Benchmark 5.1.2: Average trajectory prediction.}
		\label{fig:smalllinear_reduced_average}
	\end{center}
\end{figure}

%A $10$-model ensemble is used, and each network inputs a $10$-step
%memory period and has $3$ hidden layers each with $10$ hidden
%nodes. The $10$-step recurrent loss is trained for $10^4$ epochs with
%a learning rate of $10^{-4}$.
Figure \ref{fig:smalllinear_reduced_example} shows an example
trajectory, and Figure \ref{fig:smalllinear_reduced_average} shows the
average $\ell_2$ error over $10^2$ test trajectories using initial
conditions uniformly sampled in $\Omega_X$. 

\subsection{Oscillatory linear system}

Next we consider a slightly more complex $2\times 2$ linear system:
\begin{equation}
\frac{d\mathbf{y}}{dt} = A\mathbf{y},\quad A=\begin{bmatrix} -\alpha^{-1} & 1 \\ -1 & -\alpha^{-1} \end{bmatrix} \text{ with } \alpha=64.
\end{equation}
Unlike the examples above, which tend to steady state monotonically after some time, the following system tends to steady state in an oscillatory manner with the size of the oscillations governed by the parameter $\alpha$.

\subsubsection{Fully observed case}

The key parameters are:
\begin{itemize}
    \item $\Delta=0.01$.
  \item $\Omega_X=[0,2]^2$.
  \item $\Nt=10^4$.
    \item $L=200$.
    \item $N_{burst}=5$
      \item 
        $N=n_{burst}\cdot \Nt = 5\times 10^4$.
                \item $\nm=0$
          \item $K=10$.
        \item DNN structure: 3 hidden layers with 10 nodes
          per layer.
        \item Learning rate is $10^{-4}$; number of epochs is $10^4$.
          \item $n_{model} = 10$.
      \end{itemize}
      
\begin{figure}[htbp]
	\begin{center}
		\includegraphics[width=0.8\textwidth]{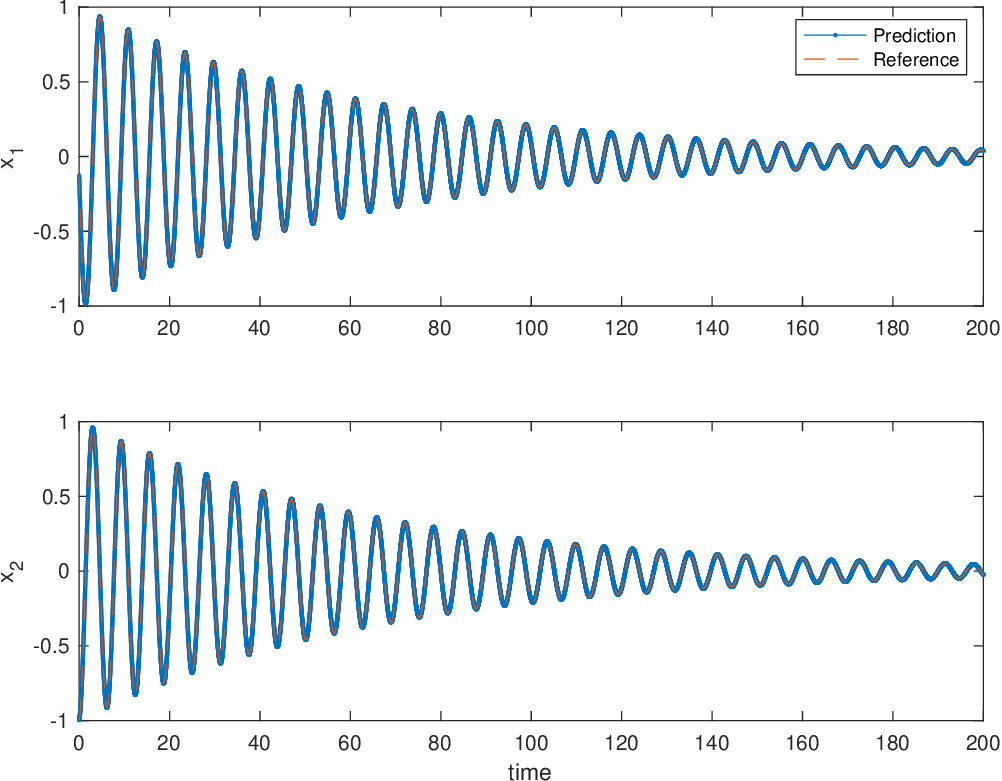}
		\caption{Benchmark 5.2.1: Example trajectory prediction.}
		\label{fig:osclinear_full_example}
	\end{center}
\end{figure}

\begin{figure}[htbp]
	\begin{center}
		\includegraphics[width=0.8\textwidth]{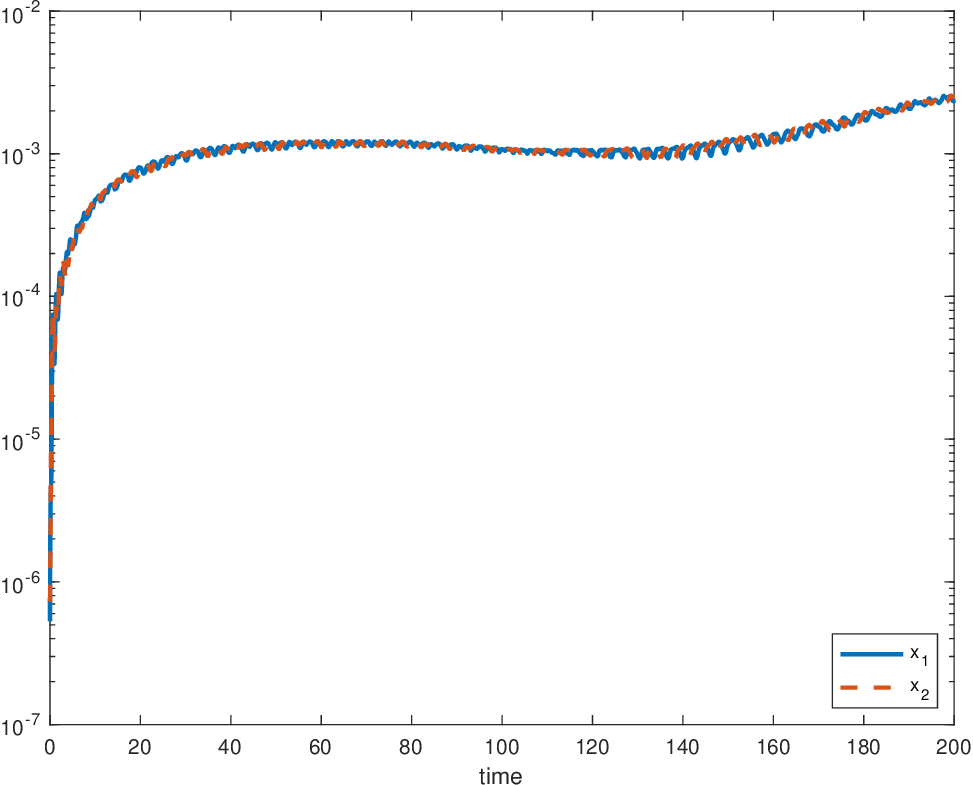}
		\caption{Benchmark 5.2.1: Average trajectory prediction.}
		\label{fig:osclinear_full_average}
	\end{center}
\end{figure}

Figure \ref{fig:osclinear_full_example} shows an example trajectory, and Figure \ref{fig:osclinear_full_average} shows the average $\ell_2$ error over $10^2$ test trajectories with initial conditions uniformly sampled from $\Omega_X$.

\subsubsection{Partially observed case}

We set the observable to be $x = y_1$. The
training data set \eqref{final_data} therefore contains only the trajectory data
of $y_1$.
The memory step, after
testing, is chosen to be $\nm=10$. The rest of the key parameters are
the same as in the previous example:
\begin{itemize}
    \item $\Delta=0.01$.
  \item $\Omega_X=[0,2]^2$.
  \item $\Nt=10^4$.
    \item $L=200$.
    \item $N_{burst}=5$
      \item 
        $N=n_{burst}\cdot \Nt = 5\times 10^4$.
                \item $\nm=10$
          \item $K=10$.
        \item DNN structure: 3 hidden layers with 10 nodes
          per layer.
        \item Learning rate is $10^{-4}$; number of epochs is $10^4$.
          \item $n_{model} = 10$.
      \end{itemize}

\begin{figure}[htbp]
	\begin{center}
		\includegraphics[width=0.8\textwidth]{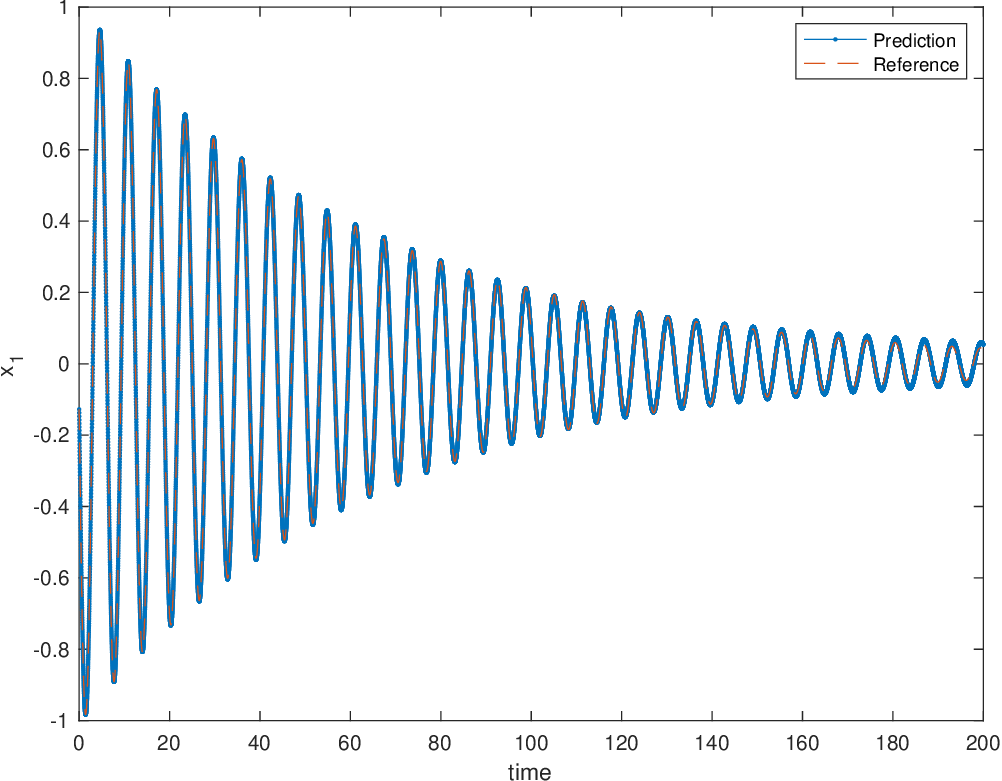}
		\caption{Benchmark 5.2.2: Example trajectory prediction.}
		\label{fig:osclinear_reduced_example}
	\end{center}
\end{figure}

\begin{figure}[htbp]
	\begin{center}
		\includegraphics[width=0.8\textwidth]{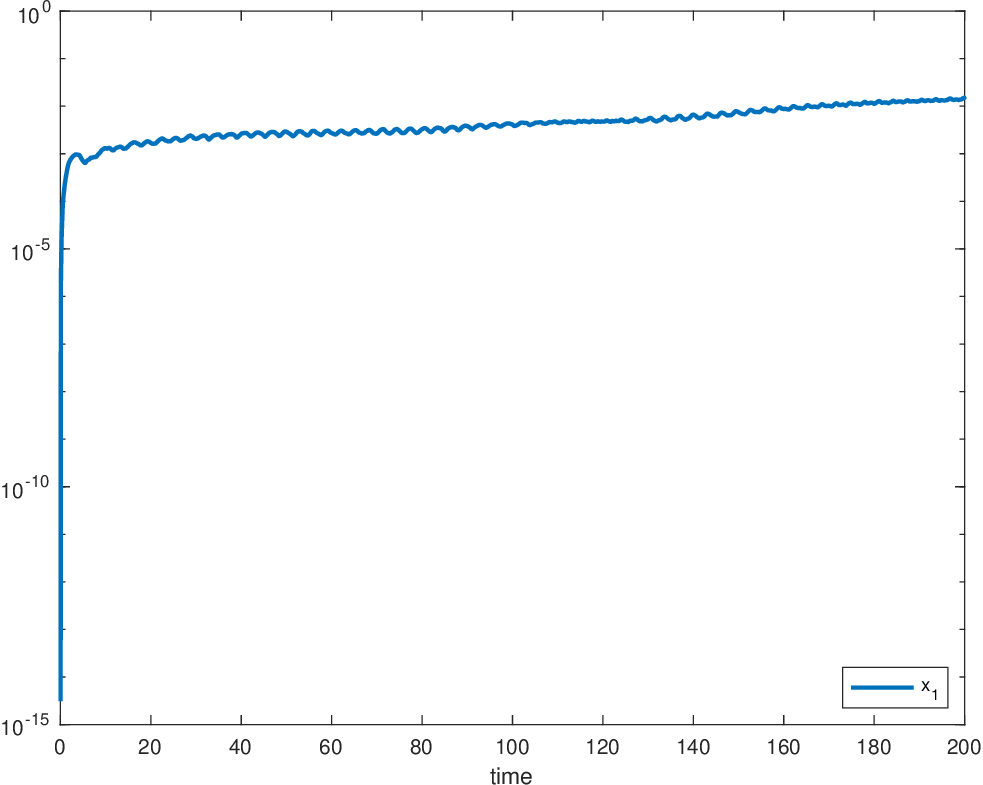}
		\caption{Benchmark 5.2.2: Average trajectory prediction.}
		\label{fig:osclinear_reduced_average}
	\end{center}
\end{figure}

Figure \ref{fig:osclinear_reduced_example} shows an example trajectory, and Figure \ref{fig:osclinear_reduced_average} shows the average $\ell_2$ error over $10^2$ test trajectories with initial conditions uniformly sampled from $\Omega_X$.

\subsection{Oscillatory nonlinear system}

The first nonlinear system we consider is a $2$-dimensional damped pendulum system:
\begin{align*}
\frac{dy_1}{dt} &= y_2 \\
\frac{dy_2}{dt} &= -\alpha y_2-\beta \sin y_1
\end{align*}
with $\alpha=10^{-1}$ and $\beta=9.80665$. The system exhibits oscillatory behavior as it decays to zero.

\subsubsection{Fully observed case}

The key parameters are:
\begin{itemize}
    \item $\Delta=0.01$.
  \item $\Omega_X=[-\pi/2,\pi/2]\times[-\pi,\pi]$.
  \item $\Nt=10^4$.
    \item $L=1000$.
    \item $N_{burst}=5$
      \item 
        $N=n_{burst}\cdot \Nt = 5\times 10^4$.
                \item $\nm=0$
          \item $K=10$.
        \item DNN structure: 3 hidden layers with 10 nodes
          per layer.
        \item Learning rate is $10^{-4}$; number of epochs is $10^4$.
          \item $n_{model} = 10$.
      \end{itemize}

\begin{figure}[htbp]
	\begin{center}
		\includegraphics[width=0.8\textwidth]{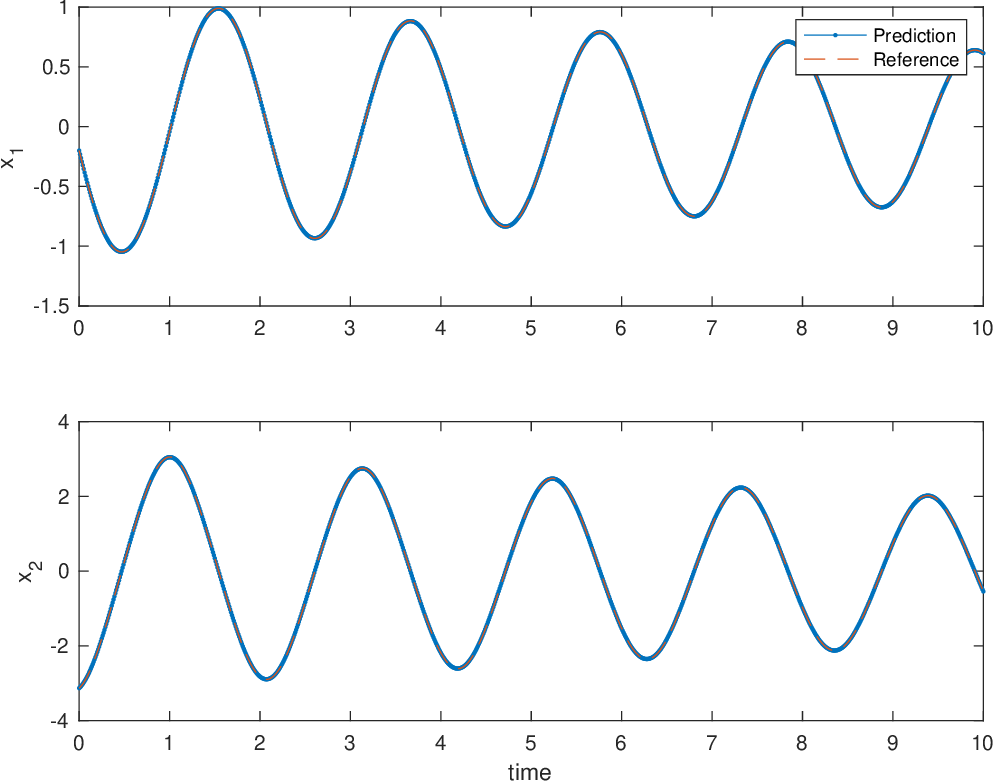}
		\caption{Benchmark 5.3.1: Example trajectory prediction.}
		\label{fig:oscnonlinear_full_example}
	\end{center}
\end{figure}

\begin{figure}[htbp]
	\begin{center}
		\includegraphics[width=0.8\textwidth]{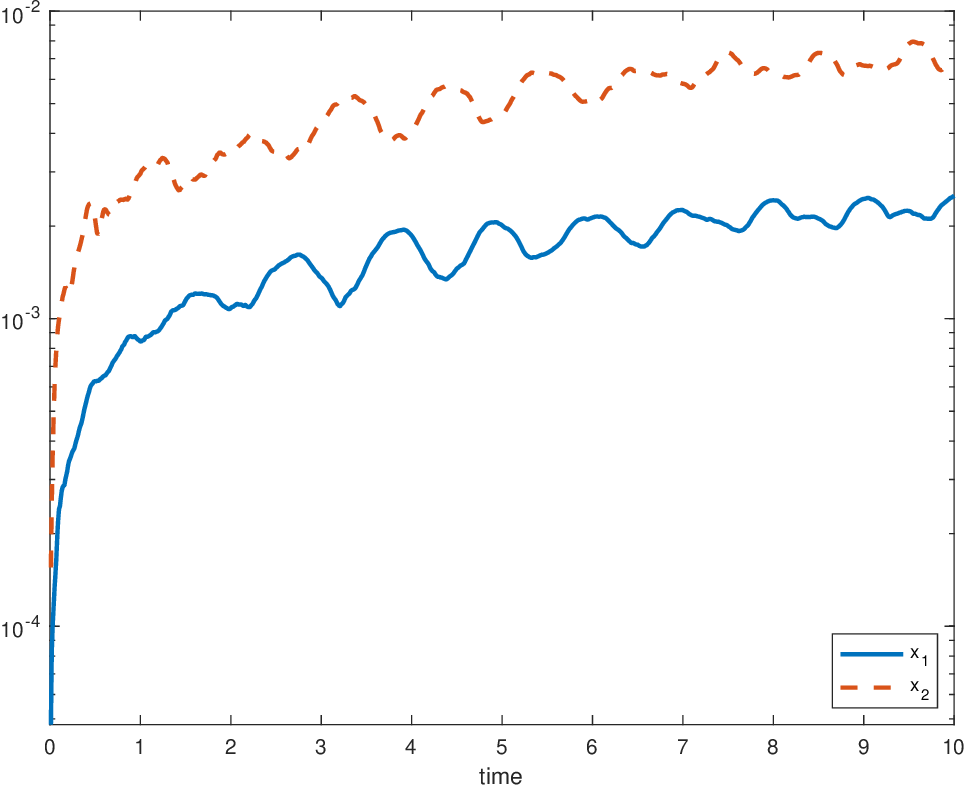}
		\caption{Benchmark 5.3.1: Average trajectory prediction.}
		\label{fig:oscnonlinear_full_average}
	\end{center}
\end{figure}

Figure \ref{fig:oscnonlinear_full_example} shows an example trajectory, and Figure \ref{fig:oscnonlinear_full_average} shows the average $\ell_2$ error over $10^2$ test trajectories with initial conditions uniformly sampled from $\Omega_X$.

\subsubsection{Fully observed case with random parameters}

In the next benchmark, instead of $\alpha$ fixed we choose $\alpha$ uniformly at random from $[0,0.2]$ for each training trajectory. Note that $\alpha$ is completely unknown in the training process, and the network relies on memory step $n_M=10$ to learn the appropriate behavior for the entire parameter domain. The rest of the key parameters are the same as above:
\begin{itemize}
    \item $\Delta=0.01$.
  \item $\Omega_X=[-\pi/2,\pi/2]\times[-\pi,\pi]$.
  \item $\Nt=10^4$.
    \item $L=1000$.
    \item $N_{burst}=5$
      \item 
        $N=n_{burst}\cdot \Nt = 5\times 10^4$.
                \item $\nm=10$
          \item $K=10$.
        \item DNN structure: 3 hidden layers with 10 nodes
          per layer.
        \item Learning rate is $10^{-4}$; number of epochs is $10^4$.
          \item $n_{model} = 10$.
      \end{itemize}

\begin{figure}[htbp]
	\begin{center}
		\includegraphics[width=0.8\textwidth]{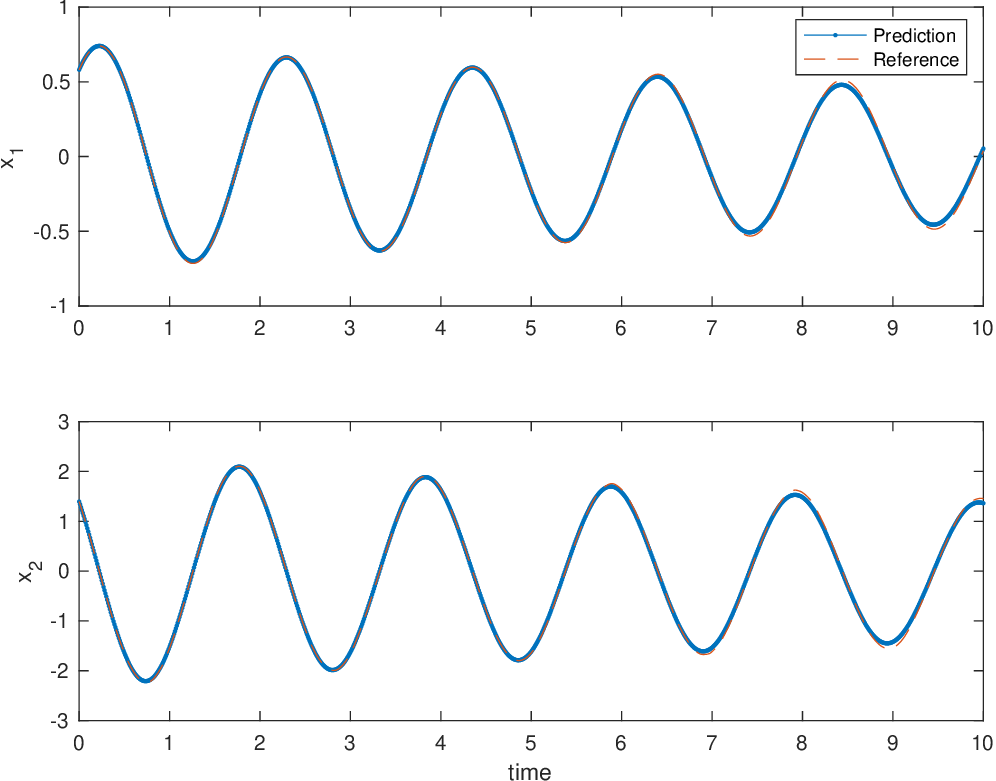}
		\caption{Benchmark 5.3.2: Example trajectory prediction.}
		\label{fig:oscnonlinear_fullrand_example}
	\end{center}
\end{figure}

\begin{figure}[htbp]
	\begin{center}
		\includegraphics[width=0.8\textwidth]{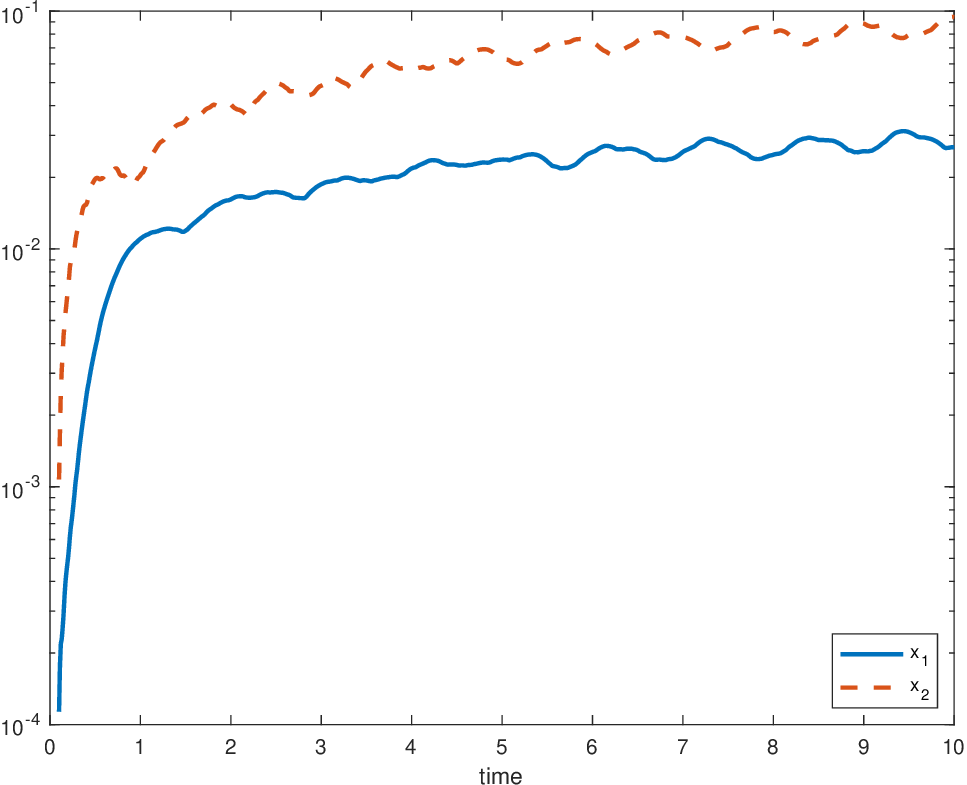}
		\caption{Benchmark 5.3.2: Average trajectory prediction.}
		\label{fig:oscnonlinear_fullrand_average}
	\end{center}
\end{figure}

Figure \ref{fig:oscnonlinear_fullrand_example} shows an example trajectory, and Figure \ref{fig:oscnonlinear_fullrand_average} shows the average $\ell_2$ error over $10^2$ test trajectories with initial conditions uniformly sampled from $\Omega_X$.

\subsubsection{Partially observed case}

Finally, we model the reduced system after observing only the first variable $x=y_1$. Key parameters are identical to the fully observed case with random parameters:

\begin{itemize}
    \item $\Delta=0.01$.
  \item $\Omega_X=[-\pi/2,\pi/2]\times[-\pi,\pi]$.
  \item $\Nt=10^4$.
    \item $L=1000$.
    \item $N_{burst}=5$
      \item 
        $N=n_{burst}\cdot \Nt = 5\times 10^4$.
                \item $\nm=10$
          \item $K=10$.
        \item DNN structure: 3 hidden layers with 10 nodes
          per layer.
        \item Learning rate is $10^{-4}$; number of epochs is $10^4$.
          \item $n_{model} = 10$.
      \end{itemize}

\begin{figure}[htbp]
	\begin{center}
		\includegraphics[width=0.8\textwidth]{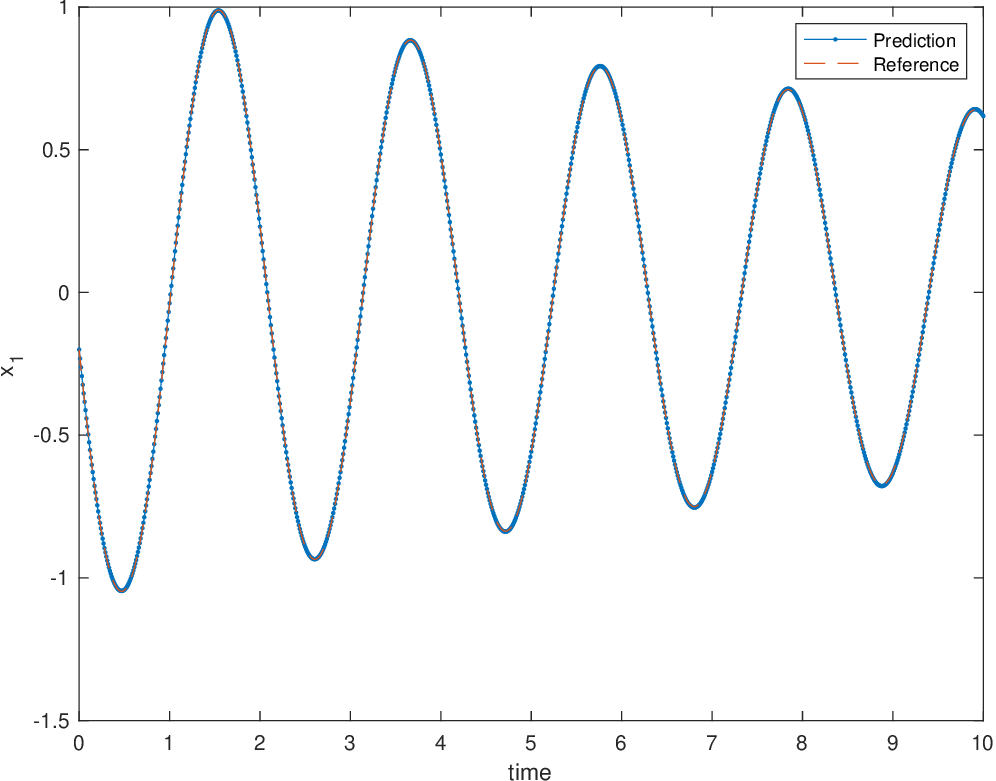}
		\caption{Benchmark 5.3.3: Example trajectory prediction.}
		\label{fig:oscnonlinear_reduced_example}
	\end{center}
\end{figure}

\begin{figure}[htbp]
	\begin{center}
		\includegraphics[width=0.8\textwidth]{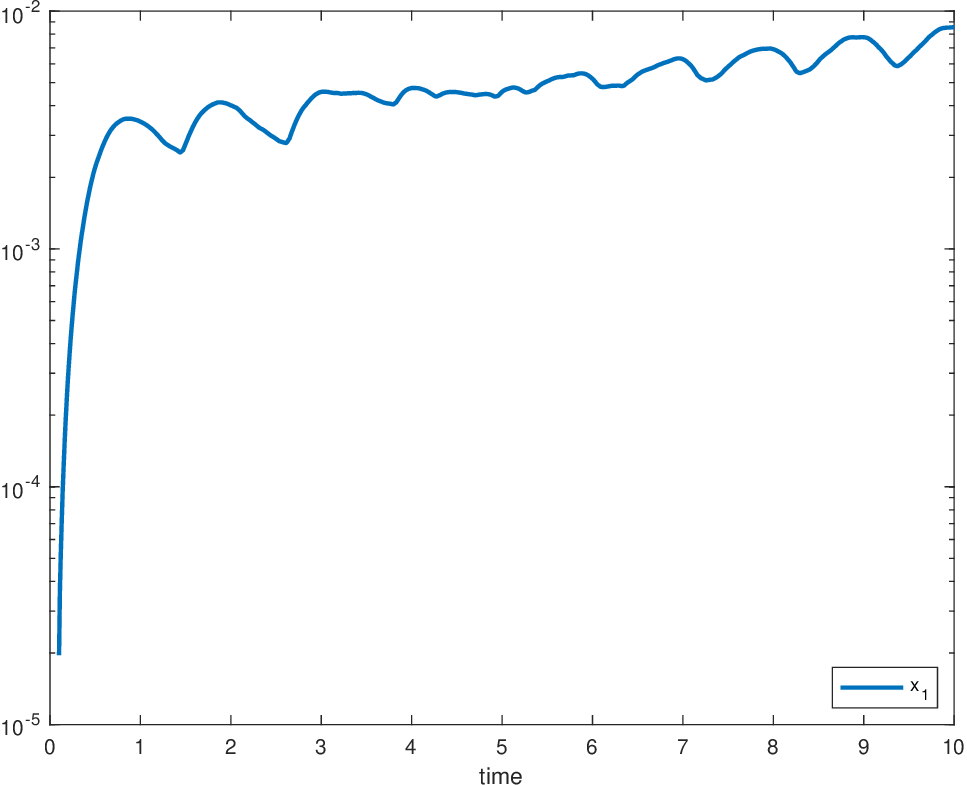}
		\caption{Benchmark 5.3.3: Average trajectory prediction.}
		\label{fig:oscnonlinear_reduced_average}
	\end{center}
\end{figure}

Figure \ref{fig:oscnonlinear_reduced_example} shows an example trajectory, and Figure \ref{fig:oscnonlinear_reduced_average} shows the average $\ell_2$ error over $10^2$ test trajectories with initial conditions uniformly sampled from $\Omega_X$.

\subsection{Chaotic nonlinear system}

A typical example in learning of dynamical systems is the Lorenz '63 system:
\begin{align*}
\frac{dx}{dt} &= \sigma(y-x) \\ 
\frac{dy}{dt} &= x(\rho-z)-y \\
\frac{dz}{dt} &= xy-\beta z
\end{align*}
with $\sigma=10$, $\rho=28$, and $\beta=8/3$. This combination of parameters causes solutions to become chaotic, arbitrarily switching between two attractors in a butterfly shape. While here we focus on three benchmarking examples for readers to reproduce, in \cite{churchill2022deep} the authors more exhaustively examine flow map learning for chaotic systems and show that the proposed approach models chaotic systems well under a variety of different metrics.

\subsubsection{Fully observed case}

The key parameters are:
\begin{itemize}
    \item $\Delta=0.01$.
  \item $\Omega_X=[-\pi/2,\pi/2]^3$.
  \item $\Nt=10^4$.
    \item $L=10,000$.
    \item $N_{burst}=5$
      \item 
        $N=n_{burst}\cdot \Nt = 5\times 10^4$.
                \item $\nm=0$
          \item $K=10$.
        \item DNN structure: 3 hidden layers with 30 nodes
          per layer.
        \item Learning rate is $10^{-4}$; number of epochs is $10^4$.
          \item $n_{model} = 10$.
      \end{itemize}

We note that using long training trajectories (large $L$) makes it more likely to select bursts from a time period when the system has fully settled into its chaotic behavior.

\begin{figure}[htbp]
	\begin{center}
		\includegraphics[width=0.8\textwidth]{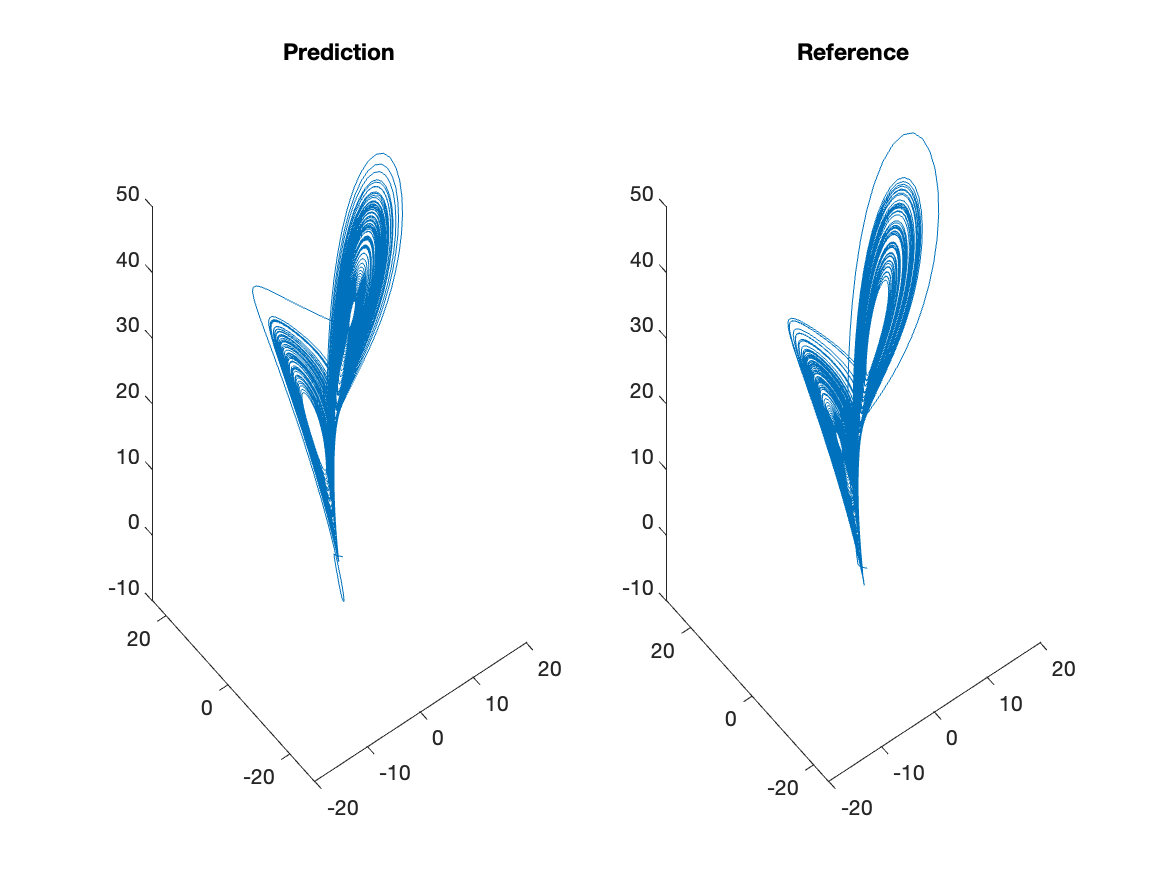}
		\caption{Benchmark 5.4.1: Example trajectory prediction.}
		\label{fig:lorenz_full_example}
	\end{center}
\end{figure}

\begin{figure}[htbp]
	\begin{center}
		\includegraphics[width=0.8\textwidth]{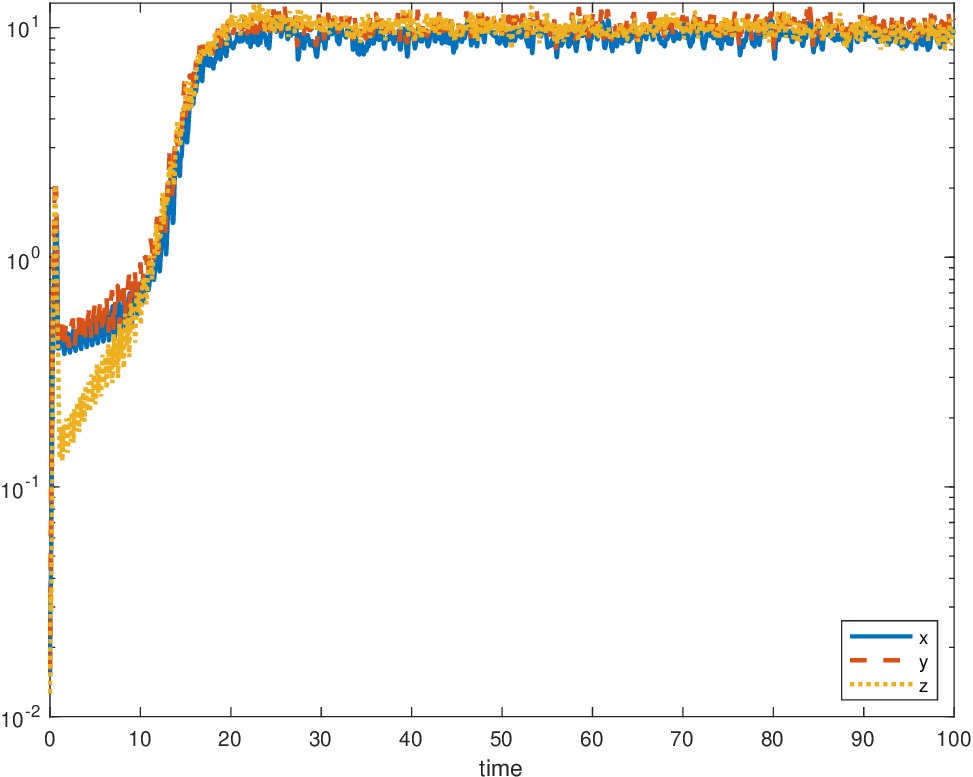}
		\caption{Benchmark 5.4.1: Average trajectory prediction.}
		\label{fig:lorenz_full_average}
	\end{center}
\end{figure}

Figure \ref{fig:lorenz_full_example} shows an example trajectory, and Figure \ref{fig:lorenz_full_average} shows the average $\ell_2$ error over $10^2$ test trajectories with initial conditions uniformly sampled from $\Omega_X$.

\subsubsection{Partially observed case of $x$ and $y$}

As another benchmark, we observe the $x$ and $y$ variables and let $z$ be unobserved. The key parameters are:
\begin{itemize}
    \item $\Delta=0.01$.
  \item $\Omega_X=[-\pi/2,\pi/2]^3$.
  \item $\Nt=10^4$.
    \item $L=10,000$.
    \item $N_{burst}=5$
      \item 
        $N=n_{burst}\cdot \Nt = 5\times 10^4$.
                \item $\nm=10$
          \item $K=10$.
        \item DNN structure: 3 hidden layers with 30 nodes
          per layer.
        \item Learning rate is $10^{-4}$; number of epochs is $10^4$.
          \item $n_{model} = 10$.
      \end{itemize}

\begin{figure}[htbp]
	\begin{center}
		\includegraphics[width=0.8\textwidth]{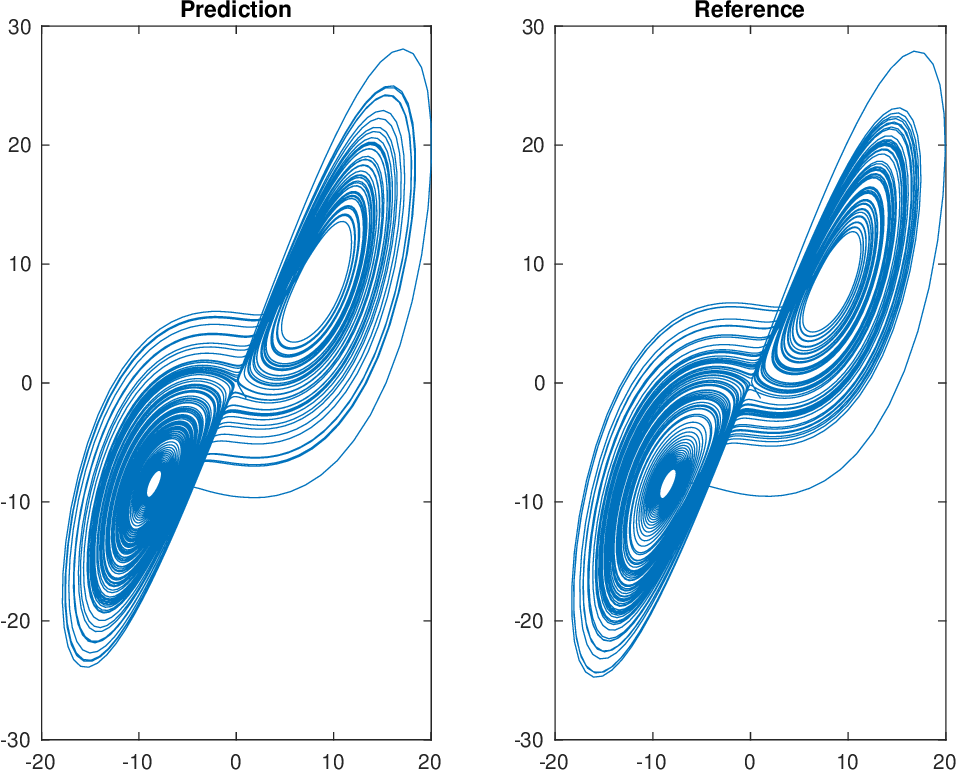}
		\caption{Benchmark 5.4.2: Example trajectory prediction.}
		\label{fig:lorenz_2d_example}
	\end{center}
\end{figure}

\begin{figure}[htbp]
	\begin{center}
		\includegraphics[width=0.8\textwidth]{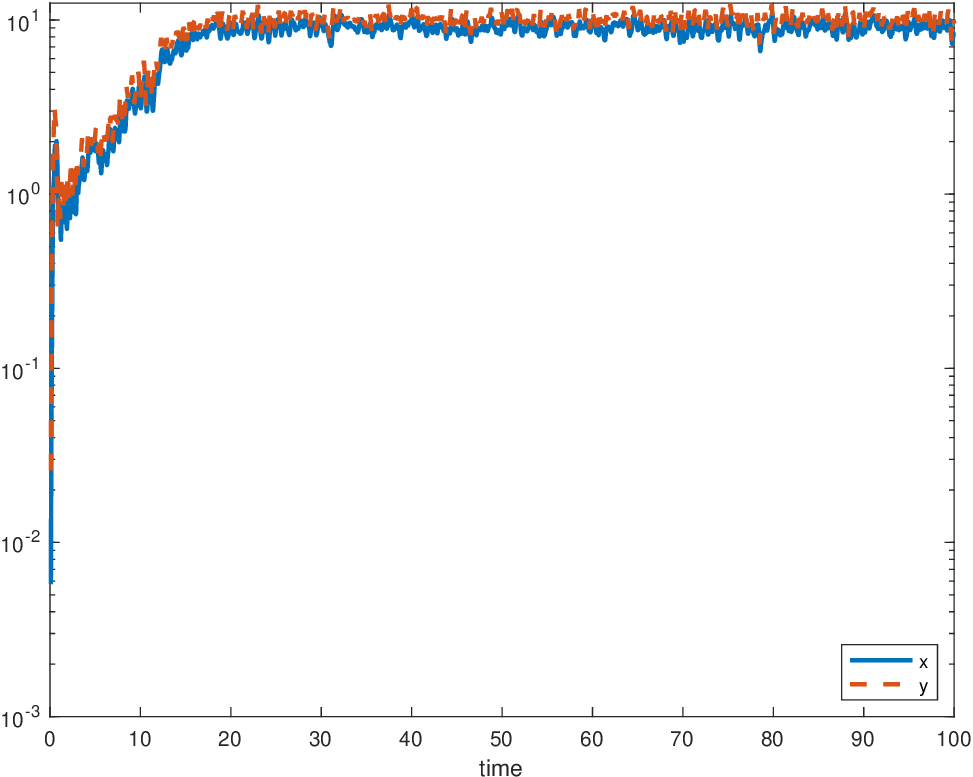}
		\caption{Benchmark 5.4.2: Average trajectory prediction.}
		\label{fig:lorenz_2d_average}
	\end{center}
\end{figure}

Figure \ref{fig:lorenz_2d_example} shows an example trajectory, and Figure \ref{fig:lorenz_2d_average} shows the average $\ell_2$ error over $10^2$ test trajectories with initial conditions uniformly sampled from $\Omega_X$.

\subsubsection{Partially observed case of $x$ only}

Finally, we consider observing only $x$ and leaving $y$ and $z$ unobserved. The key parameters are:
\begin{itemize}
    \item $\Delta=0.01$.
  \item $\Omega_X=[-\pi/2,\pi/2]^3$.
  \item $\Nt=10^4$.
    \item $L=10,000$.
    \item $N_{burst}=5$
      \item 
        $N=n_{burst}\cdot \Nt = 5\times 10^4$.
                \item $\nm=10$
          \item $K=10$.
        \item DNN structure: 3 hidden layers with 30 nodes
          per layer.
        \item Learning rate is $10^{-4}$; number of epochs is $10^4$.
          \item $n_{model} = 10$.
      \end{itemize}
   
\begin{figure}[htbp]
	\begin{center}
		\includegraphics[width=0.8\textwidth]{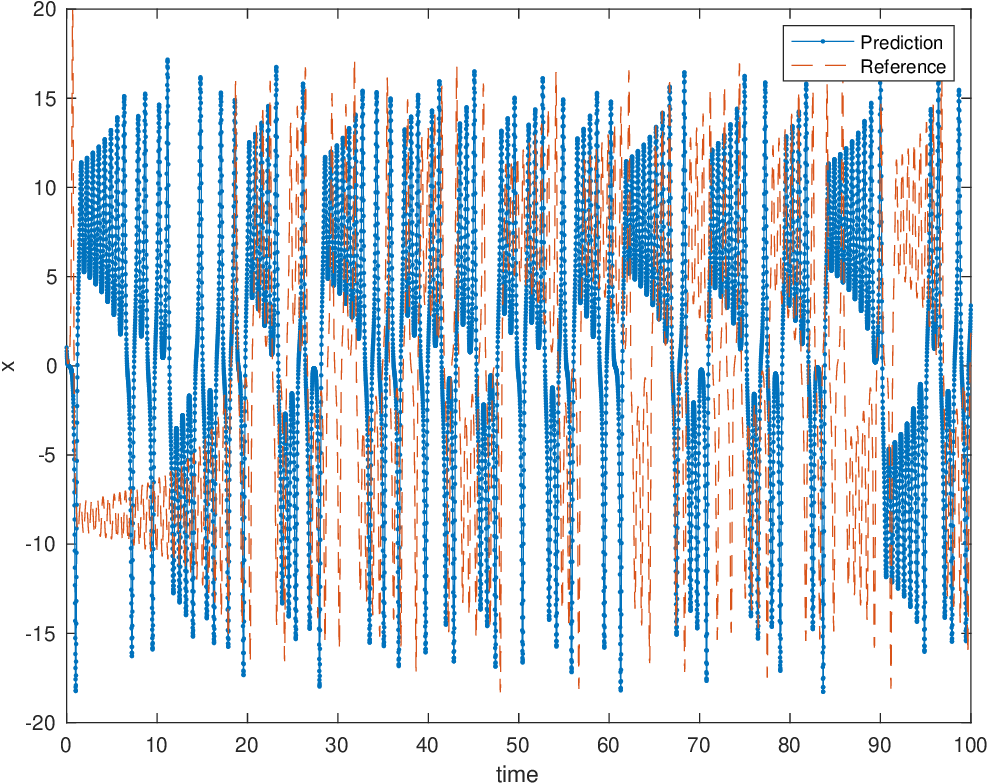}
		\caption{Benchmark 5.4.3: Example trajectory prediction.}
		\label{fig:lorenz_1d_example}
	\end{center}
\end{figure}

\begin{figure}[htbp]
	\begin{center}
		\includegraphics[width=0.8\textwidth]{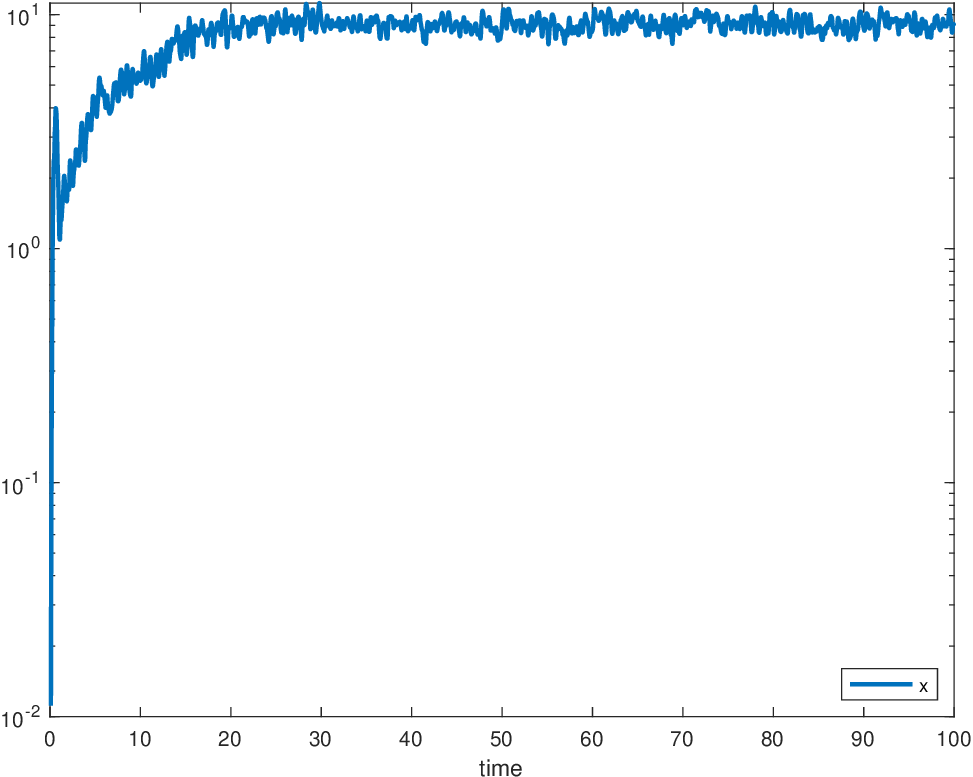}
		\caption{Benchmark 5.4.3: Average trajectory prediction.}
		\label{fig:lorenz_1d_average}
	\end{center}
\end{figure}

Figure \ref{fig:lorenz_1d_example} shows an example trajectory, and Figure \ref{fig:lorenz_1d_average} shows the average $\ell_2$ error over $10^2$ test trajectories with initial conditions uniformly sampled from $\Omega_X$.

\subsection{Large linear system}

The final benchmark is a larger $20\times 20$ linear system:
\begin{equation}
\frac{d\mathbf{y}}{dt} = A\mathbf{y}, \quad A = \begin{bmatrix} \Sigma_{11} & I+\Sigma_{12} \\ -(I+\Sigma_{21}) & -\Sigma_{22} \end{bmatrix},
\end{equation}
where $I$ is the $10\times10$ identity matrix, and the fixed constant $\Sigma$ matrices are given in the Appendix.

\subsubsection{Partially observed case}

We model the reduced system after observing only the first $10$ variables $y_1,\ldots,y_{10}$. The key parameters are:
\begin{itemize}
    \item $\Delta=0.02$.
  \item $\Omega_X=[-2,2]^{20}$.
  \item $\Nt=10^4$.
    \item $L=100$.
    \item $N_{burst}=5$
      \item 
        $N=n_{burst}\cdot \Nt = 5\times 10^4$.
                \item $\nm=30$
          \item $K=10$.
        \item DNN structure: 3 hidden layers with 100 nodes
          per layer.
        \item Learning rate is $10^{-4}$; number of epochs is $10^4$.
          \item $n_{model} = 10$.
      \end{itemize}
   
\begin{figure}[htbp]
	\begin{center}
		\includegraphics[width=0.8\textwidth]{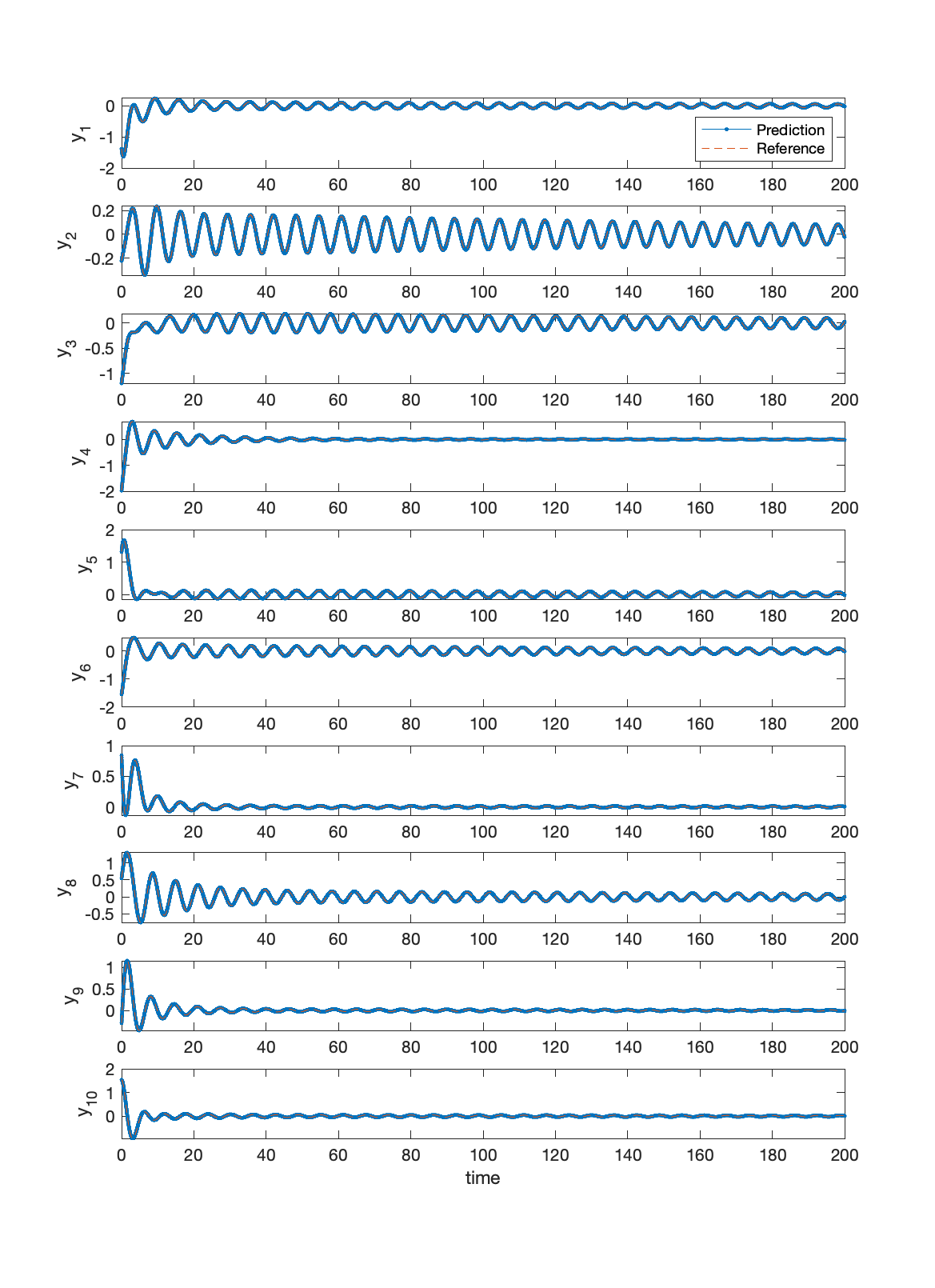}
		\caption{Benchmark 5.5.1: Example trajectory prediction.}
		\label{fig:largelinear_example}
	\end{center}
\end{figure}

\begin{figure}[htbp]
	\begin{center}
		\includegraphics[width=0.8\textwidth]{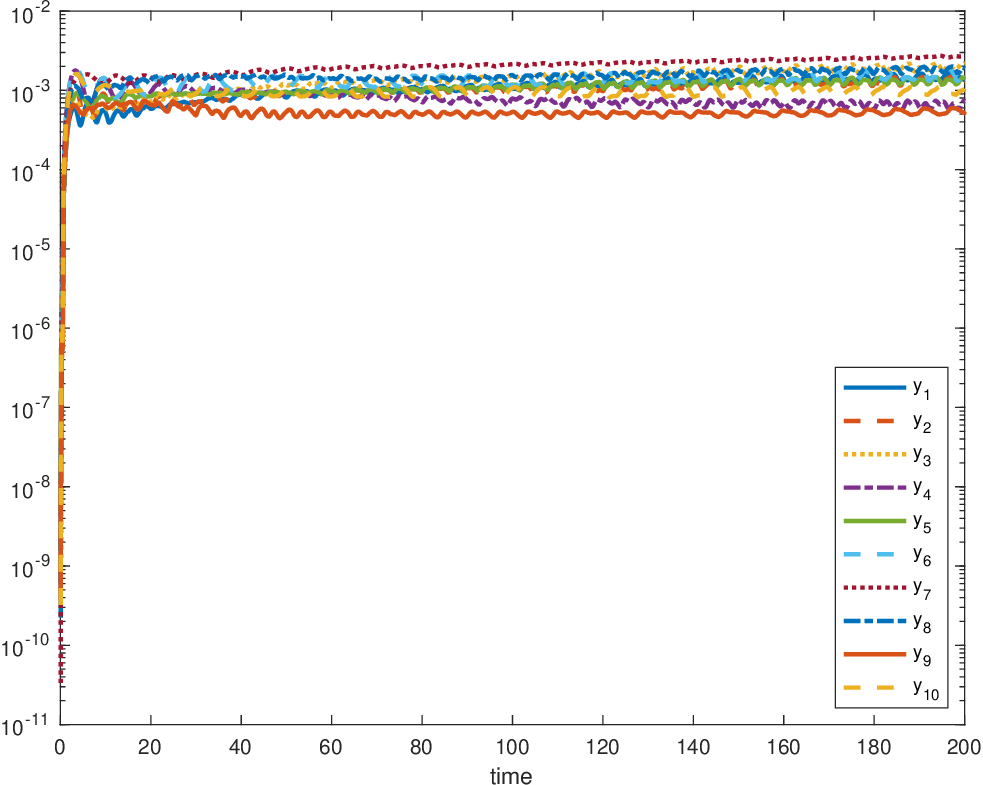}
		\caption{Benchmark 5.5.1: Average trajectory prediction.}
		\label{fig:largelinear_average}
	\end{center}
\end{figure}

Figure \ref{fig:largelinear_example} shows an example trajectory, and Figure \ref{fig:largelinear_average} shows the average $\ell_2$ error over $10^2$ test trajectories with initial conditions uniformly sampled from $\Omega_X$.

%\end{document}

%% file: Appendix.tex
\section{Appendix} \label{sec:appendix}

\begin{equation}
\Sigma_{11} = 10^{-3}\times \begin{bmatrix} -6.09 & 5.79 & -0.945 & -12.1 & 9.38 & -12.4 & 4.92 & 3.71 & 1.17 & 4.73\\
    -9.57 & -8.88 & -12.1 & -12.9 & 5.11 & 26.5 & -7.33 & -8.01 & -21.6 & -10.2\\
    -0.733 & 6.2 & 10.7 &-6.06 &-7.07& -1.7& -16.4& 6.69& -1.59& 7.69\\
    7.83& 12.5& 5.77& -14.9& -17.8 &-1.01& -4.05& -15 &-6.61& -4.94\\
    8.1& 4.13& 4.21& 23.3& -4.63& 1.77 &-14.9 &17.9& -17.1& -8.19\\
    -7.68& 6.98 &27.6& 19 &20.9& 12.2 &15.6 &-11.2 &-3.56 &-2.47\\
    -14.9 &-5.73& -19.7 &-8.77& -9.17 &-2.95& -9.48& -2.95 &5.43& 15.4\\
    -1.84& 2.05 &-1.98 &3.83 &-4.06 &7.72 &4.04 &-13.7 &20.3 &0.509\\
    12.1 &19.7& -14.3& 12.6& -4.67& 9.72 &5.87& 0.664& -10.8 &-18.2\\
    3.07 &3.65& 3.88& 7.44& 12.7 &13.5& -6.66& -23.9 &-11.7 &16.6\end{bmatrix}
\end{equation}

\begin{equation}
\Sigma_{12} = 10^{-3}\times \begin{bmatrix} 
11.7 &-12.3& -8.87& -6.86& -9.6& 11& 25.6 &-0.155& 17.8& -10.9\\
    12.9& 3.28 &2.84& 3.35& 16.6& 5.96& 6.99 &-20.2& 8.37& -8.87\\
    -0.154 &-16.5& 12.1& 0.381& 11.2& -2.59& 12.8 &3.32 &-10.9& -3.81\\
    6.49 &15.8& -0.273& 9.05 &-3.15 &0.976 &-7.35 &0.889 &6.41 &15.6\\
    4.86& -1.52 &0.118& 17.8& -5.08& -4.96 &-2.89& 3& 22.4 &16.4\\
    7.83& -9.66& -2.09 &5.97& 3.97& 19.2& 4.03& -15.3& -8.5 &-15.8\\
    -4.61& -4.98 &17& -14& -17.5& 0.104 &-27.5& 10.9& -17.9& -5.9\\
    3.88 &14 &-2.63 &-7.27& -21& -0.403& -2.18& -22 &2.01& -2.45\\
    14.4& -4.65& -8.67& -23.2& -2.73 &9.58& -13.9& 0.415 &10.3 &17.5\\
    -16.8 &8.18 &-12.3 &14.2& -18.4& -10.2& -11.4& -1.99& -2.65 &-2.34
\end{bmatrix}
\end{equation}

\begin{equation}
\Sigma_{21} = 10^{-3}\times \begin{bmatrix} 
-3.51 &-4.91& -4.51& -15.8 &-12 &-5.72& -9.52 &-14.3 &0.745 &-11.8\\
    1.8& 2.07& 8.78& 5.3& -5.25& 5.7& 0.0957 &9.77 &2.17& 12.8\\
    -9.87& 5.19& 0.884& 2.59& -7.95& 5.56 &6.41& 16.4 &15.6& 14.3\\
    10.4 &7.14 &15.5& -6.6 &5.33 &-3.37& 2.8 &-9.61& 8 &-16.8\\
    15.5& 19.6& -1.1& 0.6& 8.38 &7.62& 3.43& 1.28 &10.3& -4.76\\
    0.119 &-9.43& -6.6 &-9.99 &-10.5 &17.8 &13.5 &-6.63 &-0.566 &-1.81\\
    -6.77& -1.42& 7.46 &3.32& 11.7& 1.3 &-6.21 &6.9 &3.89& 18.9\\
    2.93 &15.1 &-4.65 &11.1 &9.13 &-9.58 &-7.04& 6.88& -4.07& 10.2\\
    -6.02& 14 &-5.91& -4.92 &0.851& 0.652 &-2.57 &0.835& -5.14 &10.6\\
    1.41& 5.8 &-2.31& 6.17 &13.3& 3.57 &15.9 &-0.753& -0.818& -10.3
\end{bmatrix}
\end{equation}

\begin{equation}
\Sigma_{22} = 10^{-3}\times \begin{bmatrix} 
1500 &124& 814& -104 &-179& -223 &-731 &-189& -400& 242\\
     124 &836 &679 &277 &197 &-515 &-52.1& -273 &101 &301\\
     814& 679& 1500 &651 &755& -605 &-379 &-546 &-225& 223\\
     -104 &277 &651& 1960 &720 &-782 &-299 &-775 &-180 &506\\
     -179 &197 &755 &720 &2290& -973 &518 &-19.1& -604 &-369\\
     -223& -515 &-605 &-782 &-973 &1290 &-400 &412& 314& -420\\
     -731 &-52.1& -379 &-299& 518& -400 &1960 &68.3 &455 &-316\\
     -189 &-273& -546 &-775 &-19.1& 412& 68.3& 576& -53.6 &-332\\
     -400& 101& -225 &-180 &-604& 314 &455 &-53.6& 1030& 265\\
     242 &301& 223 &506& -369& -420& -316 &-332 &265 &1090
\end{bmatrix}
\end{equation}